\documentclass{article}

\PassOptionsToPackage{numbers, compress, sort&compress}{natbib}
\usepackage[main, final]{neurips_2026}

\usepackage[utf8]{inputenc}
\usepackage[T1]{fontenc}

\usepackage{hyperref}
\usepackage{url}
\usepackage{booktabs}
\usepackage{amsfonts}
\usepackage{amsmath}
\usepackage{amssymb}
\usepackage{graphicx}
\usepackage{xcolor}
\usepackage{microtype}
\usepackage{amsthm}
\usepackage{multirow}
\usepackage{caption}
\usepackage{wrapfig}
\usepackage{tablefootnote}
\usepackage{tikz}
\usetikzlibrary{arrows.meta, calc, positioning, shapes, backgrounds, fit}

\newtheorem{theorem}{Theorem}
\newtheorem*{theorem*}{Theorem}
\newtheorem{proposition}[theorem]{Proposition}
\newtheorem*{proposition*}{Proposition}

\newcommand{\DeltaH}{\Delta H}
\newcommand{\HMoE}{H_{\text{MoE}}}
\newcommand{\method}{MAEP\xspace}
\usepackage{xspace}

\title{Modality-Aware Expert Pruning for MoE-Based Multimodal Large Language Models}

\author{
  Donggeun Kim \\
  Nota Inc. \\
  \texttt{donggeun.kim@nota.ai}
}

\begin{document}

\maketitle

\begin{abstract}
Mixture-of-Experts (MoE) enables multimodal large language models
(MLLMs) to scale capacity efficiently, but expert parameters dominate
memory even though only a few experts are activated per token: they
account for over 90\% of model size in modern MoE-MLLMs (e.g., 54\,GB
out of ${\sim}$58\,GB in BF16 for InternVL3.5-30B-A3B and
Qwen3-VL-30B-A3B).
Expert pruning has emerged as an effective compression strategy, but
existing methods aggregate importance uniformly across
tokens, ignoring that the same expert may serve different roles for
vision versus language processing.
To address this, we propose \method (\underline{M}odality-\underline{A}ware
\underline{E}xpert \underline{P}runing), a training-free framework
that decomposes expert importance by modality in a single forward pass.
\method introduces (1)~Delta-H Output ($\DeltaH$), a
redistribution-aware metric measuring the MoE output change when an
expert is removed, and (2)~low-attention vision tokens as a
vision-side pruning signal that is \emph{dataset-stable}: its
expert-importance ranking remains stable across different datasets.
Text-side importance serves as a preservation signal while vision-side
importance serves as a pruning signal, and experts are ranked globally
across layers.
Experiments on three MoE-based MLLMs across multimodal and text 
benchmarks show that \method outperforms modality-agnostic baselines
on the majority of multimodal benchmarks while
remaining competitive on text-only tasks, reducing the BF16 expert
weight footprint by ${\sim}27$\,GB ($\sim$47\% of total memory) so
that the full weight footprint of a ${\sim}58$\,GB model fits within a single 40\,GB GPU.
Code is available at \url{https://github.com/nota-github/MAEP}.
\end{abstract}

\section{Introduction}
\label{sec:intro}

Recent advances in large language models (LLMs) have increasingly adopted
the Mixture-of-Experts (MoE) architecture as a key strategy for scaling
model capacity while maintaining computational
efficiency~\cite{deepseek2024v3, yang2025qwen3, jiang2024mixtral}.
By activating only a subset of experts per token through sparse routing,
MoE models achieve a favorable trade-off between total parameter count
and inference cost.
This paradigm has naturally extended to multimodal LLMs
(MLLMs)~\cite{wang2025internvl35, bai2025qwen3vl, wu2024deepseekvl2,
team2025glm45v, kimi2025vl, baidu2025ernie45, meta2025llama4},
where MoE layers handle the joint processing of vision and
language inputs.

However, while sparse MoE reduces computation by activating only a few
experts per token, all expert parameters must remain in memory regardless
of activation---making the MoE architecture inherently memory-intensive.
In models such as InternVL3.5-30B-A3B and Qwen3-VL-30B-A3B,
expert parameters account for 93--94\% of the total model size
(54\,GB out of ${\sim}$58\,GB in BF16; see Appendix~\ref{app:model_config}),
despite only 8 of 128 experts being activated per token.
This gap between memory footprint and effective utilization motivates
\textbf{expert compression}---reducing the number or size of experts
through techniques such as pruning, merging, or quantization.
For instance, pruning 50\% of experts can reduce the BF16
\emph{routed-expert weight footprint} by ${\sim}$27\,GB (47\% of
total weight memory), bringing the BF16 weight footprint of a 58\,GB model below the nominal
capacity of a single 40\,GB GPU, thereby enabling deployment in
memory-constrained settings.

Existing expert compression approaches, including merging~\cite{li2024mcsmoe, chen2024hcsmoe, li2025submoe},
pruning~\cite{lasby2025reap, bai2025diep}, and decomposition/quantization~\cite{huang2024mcmoe, li2025moesvd},
have proven effective for text-only MoE LLMs.
However, these methods aggregate importance statistics uniformly
across all input tokens without distinguishing token modality,
making their direct application to MoE-based MLLMs inadequate.
Indeed, recent analysis~\cite{huang2025modes} reveals a consistent
distributional gap between text and vision tokens in expert layers,
with each modality interacting with experts differently,
highlighting the need for a modality-aware expert compression strategy.

To this end, we propose
\textbf{\method (\underline{M}odality-\underline{A}ware
\underline{E}xpert \underline{P}runing)},
a training-free framework with two key components.
Given only a small calibration set, \method computes all importance
scores in a single forward pass without any gradient computation or
weight updates:

First, we propose \textbf{Delta-H Output} ($\DeltaH$,
where $H$ denotes the MoE layer's hidden-state output)
as the expert importance metric, defined as the squared L2 change
in this output when an expert is removed with routing redistribution.
Unlike metrics based on routing frequency or an expert's isolated
contribution, $\DeltaH$ captures the residual output change after redistributing weight to the remaining experts; 
smaller values indicate higher functional redundancy among co-activated experts.
Because $\DeltaH$ averages over a token subset, it can be computed
separately for text and vision tokens, yielding per-modality importance
scores.

Second, we find that not all vision tokens are equally reliable for
importance estimation.
Evaluating $\DeltaH$ over the full set of vision tokens yields
dataset-dependent rankings because vision content is tied to
image-specific objects, making the resulting importance estimates
domain-sensitive.
Instead, \textbf{low-attention vision tokens}---those receiving
minimal attention from text tokens---provide a substantially more
dataset-stable signal.

We evaluate \method across multiple MoE-based MLLMs, pruning ratios, and benchmarks. 
Our results suggest that modality-aware importance scoring outperforms modality-agnostic baselines, 
supporting the value of modality decomposition for expert compression in vision-language models.

\section{Related Work}
\label{sec:related}

\subsection{Expert Compression for MoE Models}
\label{sec:related_compression}

Prior work on compressing MoE-based LLMs has explored several directions.
Among methods that reduce the expert pool, two prominent strategies
are \emph{merging}, which combines similar experts~\cite{
li2024mcsmoe, chen2024hcsmoe, li2025submoe, nguyen2025camex,
nguyen2026namex, miao2025mergemoe}, and \emph{pruning}, which removes
redundant experts based on estimated importance.
Pruning methods differ mainly in how they identify redundant
experts---e.g.,
router-weighted activation norms~\cite{lasby2025reap},
differentiable non-uniform layer-wise expert
pruning~\cite{bai2025diep}, second-order atomic-expert
removal~\cite{li2025heapr}, or calibration-based search over preserved expert subsets~\cite{lu2024notallexperts}.
At the weight level, MoE-Pruner~\cite{xie2024moepruner} extends
activation-aware weight pruning to LLMs by scaling
weight-activation scores with router weights.
Complementary strategies include replacing pruned experts with
lightweight replacement modules~\cite{zhang2025mone},
neuron-level recombination~\cite{zhou2025dern}, and
quantization~\cite{huang2024mcmoe} or low-rank
decomposition~\cite{li2025moesvd, gu2025d2moe, yang2024moei2}.
However, most prior pruning and merging methods have been developed
primarily for LLMs, leaving expert-compression strategies
for MoE-MLLMs relatively underexplored.

Moreover, many importance-based criteria---particularly those derived
from activation frequency or routing
statistics~\cite{lu2024notallexperts, xie2024moepruner}---aggregate
signals uniformly over tokens without distinguishing modality, which
can obscure modality-specific expert roles in vision-language models.
Vision tokens are themselves substantially
redundant~\cite{chen2024fastv, zhang2024sparsevlm}, which raises the
concern that frequency-aggregated importance may overweight
non-essential tokens; indeed, calibrating weight pruning on all text
tokens but only a subset of visual tokens has been shown
to improve pruning quality~\cite{li2026atvpruning}.
This motivates modality-aware importance
measures that decompose expert contributions by token type rather
than aggregating uniformly across all tokens.

\subsection{Efficient MLLMs}
\label{sec:related_multimodal}

As MLLMs grow larger, their memory and compute demands increasingly
constrain real-world deployment.
Prior work has improved inference efficiency by reducing the number of
visual tokens forwarded to the language
model~\cite{chen2024fastv, zhang2024sparsevlm, luan2025adaptprune},
lowering weight precision through
quantization~\cite{li2025mbq, yu2025mquant}, and compressing
the KV cache to reduce runtime
memory~\cite{tu2024vlcache, huang2025aircache}.

As MoE architectures---which scale model capacity while keeping
per-token computation fixed---have also become common in
MLLMs~\cite{lin2024moellava, wang2025internvl35, bai2025qwen3vl},
improving efficiency for MoE-based architectures has been explored.
FastMMoE~\cite{xia2025fastmmoe} combines expert activation reduction
for visual tokens with routing-aware visual token pruning, while RoE~\cite{wu2024roe} and
SMAR~\cite{xia2025smar} optimize expert routing strategies.
MoDES~\cite{huang2025modes} is closest in motivation to our work:
it accounts for modality-dependent expert utility and
applies modality-specific thresholds to skip experts during inference.
Existing MoE-MLLM efficiency methods mainly reduce inference-time
computation through expert skipping or visual-token
pruning; the expert parameters themselves, however, persist in GPU
memory. \method targets this complementary but underexplored problem:
permanent routed-expert pruning for MoE-based MLLMs, using
modality-decomposed importance scores.
\section{Preliminaries}
\label{sec:prelim}

\subsection{Mixture-of-Experts Layers}
\label{sec:prelim_moe}

Within each transformer block, the Mixture-of-Experts (MoE) architecture
replaces the dense feed-forward network (FFN) with a set of $N$ parallel
expert networks $\{E_1, \ldots, E_N\}$ and a lightweight router
network $G$.
For an input hidden state $h \in \mathbb{R}^d$, the router applies a linear projection and a scoring function---typically softmax or
sigmoid---to produce per-expert scores
$p_i(h)$.
Only the top-$k$ highest-scoring experts are activated for each token,
and the MoE layer output is their weighted sum:
\begin{equation}
\HMoE^{(l)}(h) = \sum_{i \in \text{top-}k(p(h))} w_i \cdot E_i(h)
\end{equation}
where $w_i = p_i / \sum_{j \in \text{top-}k} p_j$ are the routing
weights.
This sparse activation enables MoE models to scale total parameters while maintaining the per-token computation cost of a much
smaller model.
\subsection{MoE-based MLLM Architecture}
\label{sec:prelim_arch}

MoE-based MLLMs use a vision encoder and a
modality projector to map images into vision tokens, which are
\textbf{concatenated with text tokens} and passed through the shared
MoE backbone.
Crucially, the router dispatches both vision and text tokens to the
\textbf{same pool of experts} based solely on the hidden state, without
explicit modality conditioning.
After the final MoE layer, the model applies an LM head
$W \in \mathbb{R}^{V \times d}$ that maps the last hidden state $h$ to
the output distribution $p(y \mid h) = \mathrm{softmax}(W h)$ over a
vocabulary of size $V$.
This shared-pool design raises a fundamental question:
\textit{is expert importance uniform across modalities, or does
it differ between text and vision tokens?}
We investigate this empirically in Section~\ref{sec:motivation}.

\section{Empirical Motivation}
\label{sec:motivation}

\subsection{Why We Need a Modality-Aware Proxy Metric}
\label{sec:motivation_proxy}

Following OBD~\cite{lecun1989obd}'s principle, which measures a component's
importance by the loss change upon removal, a direct way to score an expert is to remove
it and measure the induced change in the model's next-token
distribution.
This change can be captured by the LM-head KL divergence
$D_{\mathrm{KL}}(p_{\text{orig}} \| p_{\text{pruned}})$ between the
unpruned and pruned output distributions.
Directly measuring this KL for every expert would capture the effect on output but is prohibitively expensive: with $N$ experts
per layer and $L$ MoE layers, it requires $N \times L$ full forward
passes per calibration sample (e.g., $128 \times 48 = 6{,}144$ passes
for InternVL3.5-30B-A3B).
We therefore use \textbf{Delta-H Output} ($\DeltaH$), the MoE-output
perturbation induced by expert removal, as a lightweight proxy.
We formalize this proxy in Section~\ref{sec:deltah}.

Furthermore, a proxy aggregating uniformly over tokens is insufficient.
Prior work has observed modality-dependent expert behavior in MoE-based
MLLMs~\cite{xia2025smar,huang2025modes,xu2026routing}; we find that this extends to expert \emph{importance}: when ranked by
$\DeltaH$ on text tokens versus vision tokens, the resulting expert
importance rankings show limited agreement (Spearman $\rho$ between $-0.11$
and $0.05$ at specific layers; Appendix~\ref{app:modality_gap}). The expert removal also induces an asymmetric output-side perturbation:
the LM-head output distribution shifts up to 29× more on vision-token
positions than on text-token positions, indicating modality-asymmetric
representational sensitivity to expert removal.
These observations imply that an effective metric must be
\textbf{computationally efficient} and \textbf{modality-decomposable}.

\subsection{Vision Token Heterogeneity and Signal Stability}
\label{sec:vision_signal}

Prior work on vision token pruning~\cite{chen2024fastv,
zhang2024sparsevlm, luan2025adaptprune} has shown that not all vision
tokens contribute equally.
This suggests that modality decomposition alone is not sufficient on
the vision side: the choice of vision tokens used to estimate expert
importance also matters.
We compute the \textbf{attention score of a vision token} as the
average text-to-vision attention weight, averaged across all heads
at each layer (a single scalar per token, layer, and sample), and compare
token subsets by cross-dataset ranking stability.

Specifically, cross-dataset Spearman $\rho$ (between
ShareGPT4V~\cite{chen2023sharegpt4v} and
ChartNet~\cite{ibm2024chartnet}, averaged across layers;
Appendix~\ref{app:low_attn_analysis}) reveals a clear hierarchy:
full vision ($\approx 0.0$), high-attention vision (top $q{\approx}20\%$, $\approx 0.2$),
text ($\approx 0.5$), and low-attention vision
($\mathbf{\approx 0.6}$).
In particular, expert importance ranked by
$\DeltaH$ computed from the full or
high-attention vision tokens is sensitive to the
calibration dataset. By construction, cross-modal attention
highlights tokens relevant to the current text query---a
property that text-guided token pruning methods such as
SparseVLM~\cite{zhang2024sparsevlm} rely on. Different
calibration datasets carry different content and queries,
so high-attention selections drift across datasets.
Moreover, not all high-attention tokens are
informative: some are visual attention
sinks~\cite{kang2025visualattnsink} that attract attention
despite carrying little semantic content.
By contrast, \textbf{low-attention vision tokens} (bottom $q$-th
percentile of text-to-vision attention) yield
the most stable expert importance rankings across calibration datasets,
suggesting they reflect the model's visual processing rather
than domain-specific content.

\section{Proposed Method}
\label{sec:method}

\begin{figure}[t]
\centering
\includegraphics[width=\linewidth]{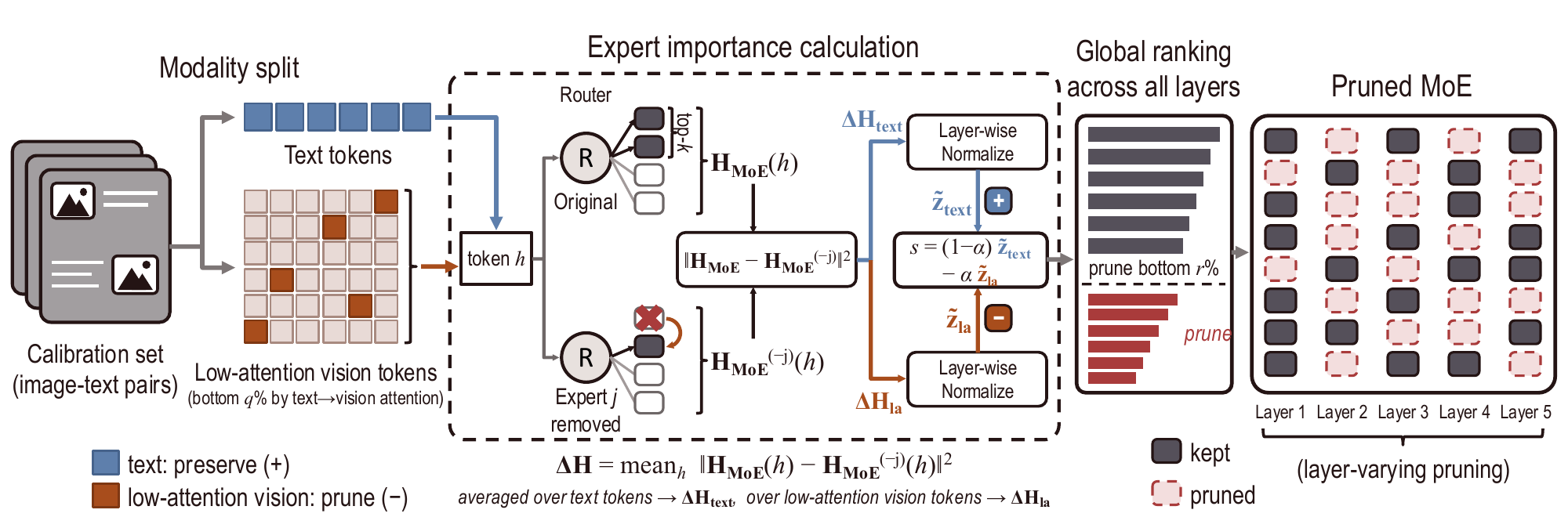}
\caption{Pipeline of \method. Text and low-attention vision tokens
are scored separately via $\DeltaH^{(l,j)}$, layer-wise normalized,                                                                                                                                         
combined with asymmetric signs---text preservation ($+$) and
low-attention vision pruning ($-$); see                                                                                                                                                                     
Eq.~\ref{eq:combined_score}---and ranked globally.}   
\label{fig:pipeline}
\end{figure}

Building on the observations in Section~\ref{sec:motivation}, this
section formally defines the $\DeltaH$ proxy and presents the
pruning framework (Figure~\ref{fig:pipeline}).
The overall pipeline is entirely training-free:
given a small calibration set\footnote{We use 1K samples from ShareGPT4V~\cite{chen2023sharegpt4v}, selected by text length; see Appendix~\ref{app:exp_settings} for details.}, the full
pipeline (importance scoring, modality decomposition, and global
pruning) completes within a single forward pass per sample.

\textbf{Scope.}
\method applies to MoE architectures with top-$k$ routing ($k\geq 2$,
required by the redistribution counterfactual; Section~\ref{sec:deltah}).
For architectures that additionally employ always-active shared experts
(e.g., DeepSeek-VL2~\cite{wu2024deepseekvl2},
Kimi-VL~\cite{kimi2025vl}), \method targets only the routed experts:
shared experts are excluded from importance scoring and pruning.
The memory savings reported in this paper refer exclusively to the
removal of routed expert parameters.

\textbf{Notation.}
We distinguish two levels of indexing:
$h \in \mathbb{R}^d$ denotes a single token's hidden state, while $x$
denotes a calibration sample (image-text pair) of $|x|$ tokens.
At the token level, $\HMoE^{(l)}(h)$ is the MoE layer output
(Section~\ref{sec:prelim_moe}), $p_i(h)$ is the router score, and
$w_i$ is the routing weight.
Sample-level and dataset-level aggregation of importance scores
is defined in Section~\ref{sec:deltah}.

\subsection{Delta-H Output: Expert Importance Metric}
\label{sec:deltah}

We define the importance of expert $j$ in layer $l$ as the expected
change in the MoE output when expert $j$ is removed:
\begin{equation}
\DeltaH^{(l,j)} = \mathbb{E}_{x \sim \mathcal{D}} \left[
\frac{1}{|x|}\sum_{h \in x}
\left\| \HMoE^{(l)}(h) - \HMoE^{(l,-j)}(h) \right\|_2^2 \right]
\label{eq:deltah}
\end{equation}
where $\HMoE^{(l,-j)}$ denotes the MoE output with expert $j$ excluded
via \textbf{router redistribution}:
expert $j$'s router logit is masked and the top-$k$ experts are
re-selected and renormalized, as in the pruned model.
Formally, the redistributed weight for each remaining expert $i \neq j$ is
$w_i^{(-j)} = w_i / (1 - w_j + w_{(k+1)})$, where $w_{(k+1)}$ is the
$(k{+}1)$-th ranked expert's router score normalized in the same way as
the top-$k$ weights, preserving the relative ratios among the remaining
experts.

Redistribution operates on renormalized top-$k$ weights, making it structurally
agnostic to the underlying scoring function.
We adopt a \textbf{fixed top-$k$ counterfactual} (the promoted
$(k{+}1)$-th expert is not evaluated and contributes no output) to avoid
an additional expert computation during calibration; this is nearly
identical to the promotion variant (mean $\rho > 0.98$ across
modalities; Appendix~\ref{app:topk_counterfactual}).

The entire computation requires only a \textbf{single forward pass} through the model:
cached top-$k$ weights and expert outputs let us evaluate all
counterfactuals analytically.
Restricting the token set to a specific modality yields per-modality
variants of $\DeltaH$, supporting fine-grained importance estimation
while capturing functional redundancy among co-activated experts.

\subsubsection{Why \texorpdfstring{$\DeltaH$}{Delta-H}: Connection to KL Divergence}
\label{sec:kl_bound}

The following bound shows that $\|\hat h - \hat h^{(-j)}\|_2^2$
(the squared $L_2$ change in the MoE output, with
$\hat h := \HMoE^{(l)}(h)$) arises naturally in a KL upper bound,
motivating $\DeltaH$ as a \emph{within-layer} ranking proxy.
It does not itself justify the cross-layer normalization,
modality decomposition, or global ranking in
Section~\ref{sec:global_pruning}; those are described and validated
separately.

\begin{proposition}
\label{prop:kl_bound}
Let $\Phi_l: \mathbb{R}^{T \times d} \to \mathbb{R}^{T \times |V|}$ map MoE layer $l$'s
outputs for a $T$-token sequence through subsequent transformer blocks and the LM head to
pre-softmax logits, on a compact activation domain (approximately
ensured by RMSNorm/LayerNorm).
For the pre-MoE hidden states $h_t \in \mathbb{R}^d$ of the sequence
at layer $l$, let $\hat h_t := \HMoE^{(l)}(h_t)$ and
$\hat h_t^{(-j)} := \HMoE^{(l,-j)}(h_t)$ denote the MoE outputs
before/after removing expert $j$, and let $p_t$ and $p_t^{(-j)}$ denote
the resulting next-token distributions at position $t$. Then
\begin{equation}
\frac{1}{T}\sum_{t=1}^{T}
D_{\textup{KL}}\!\bigl(p_t \,\|\, p_t^{(-j)}\bigr)
\leq \frac{L_{\Phi_l}^2}{4}\cdot\frac{1}{T}\sum_{t=1}^{T}
\|\hat h_t - \hat h_t^{(-j)}\|_2^2,
\end{equation}
where $L_{\Phi_l} = \|W\|_{\textup{op}}\,L_{\mathrm{norm}}\prod_{i=l+1}^{L}(1+L_{g_i})$
absorbs the LM head, the pre-LM-head normalization, and the local
Lipschitz constants of subsequent attention+FFN branches $g_i$ in
$f_i(X)=X+g_i(X)$.
\end{proposition}

The proof (Appendix~\ref{app:proof}) composes Lipschitz post-MoE blocks
with the $\tfrac{1}{2}$-smoothness of $A(z)=\log\sum_i e^{z_i}$.
Averaging over the calibration distribution yields
$\mathbb{E}_{x \sim \mathcal{D}}\!\left[\tfrac{1}{|x|}\sum_{h \in x} D_{\mathrm{KL}}\right]
\leq C_l \cdot \DeltaH^{(l,j)}$, with $C_l = L_{\Phi_l}^2 / 4$.

\textbf{Role of the bound.}
Since $C_l$ is layer-dependent but expert-independent, ranking
experts by the bound within a layer is equivalent to ranking them by
$\DeltaH$.
The bound motivates $\DeltaH$---rather than routing frequency
or gating entropy---as the quantity that emerges naturally from the
Lipschitz--softmax composition of the post-MoE network.
Since $C_l$ is a layer-shared constant, within-layer ranking
depends only on relative magnitudes of $\DeltaH$
across experts; $L_{\Phi_l}$ thus does not affect ranking
outcomes and need not be estimated.

\textbf{Empirical validation.}
Per-layer Spearman rank correlation between $\DeltaH$ and actual KL
divergence increases with depth overall for both modalities,
reaching ${>}0.8$ in deeper layers (vision) and ${\sim}0.89$ (text)
(Appendix~\ref{app:spearman_depth}).
This pattern is qualitatively consistent with the depth
dependence in Proposition~\ref{prop:kl_bound}: deeper layers admit a
tighter bound and thus a more reliable proxy.
The weaker correlation in shallow layers reflects the bound's
depth dependence in Proposition~\ref{prop:kl_bound}; in practice,
the global pruning decision aggregates across all layers via the
combined score (Eq.~\ref{eq:combined_score}).

\subsection{Low-Attention Vision Scoring}
\label{sec:low_attn}

Motivated by the observation in Section~\ref{sec:vision_signal} that
full-vision and high-attention signals are dataset-dependent, we use
low-attention vision tokens as the basis for the vision-side pruning
signal.
Per sample and per layer, tokens in the bottom $q$-th percentile ($q = 2\%$;
ablated in Section~\ref{sec:ablation_components}) of the attention
score (Section~\ref{sec:vision_signal}) are classified as
low-attention; this per-sample selection adapts to varying image
resolutions and aspect ratios.
We then compute $\DeltaH$ (Eq.~\ref{eq:deltah}) over these tokens only
(with $|x|$ rescoped to the subset cardinality), yielding the
vision-side score $\Delta H_{\text{la}}^{(l,j)}$
(``la'' denoting low-attention; same subscript convention used for
$\tilde{z}_{\text{la}}$ in Eq.~\ref{eq:combined_score}).
A high value indicates that expert $j$ is heavily engaged with vision
tokens that receive little text-to-vision attention---i.e., experts
that specialize in non-salient background visual content.
\textbf{In contrast to text-side $\DeltaH$, where a high value marks
an essential expert to preserve, a high
$\Delta H_{\text{la}}^{(l,j)}$ identifies experts that allocate
capacity to less informative tokens and are therefore favorable pruning
candidates} (verified at the model output in Appendix~\ref{app:kl_sign});
we make this asymmetric usage explicit in
Section~\ref{sec:global_pruning}.
This signal is also more dataset-stable than full or high-attention
alternatives (Section~\ref{sec:vision_signal}), since low-attention
tokens reflect non-salient background content rather than domain-specific
features (Appendix~\ref{app:spatial_stats}).

\subsection{Combined Scoring and Global Pruning}
\label{sec:global_pruning}

The two signals play \textbf{asymmetric roles}:
text-side $\DeltaH$ acts as a \textit{preservation signal}
(high $\to$ retain), while low-attention vision-side $\DeltaH$ acts as
a \textit{pruning signal} (high $\to$ remove).
Since raw $\DeltaH$ values follow a heavy-tailed distribution and vary
in magnitude across layers by up to $10^3$--$10^4\times$
(Appendix~\ref{app:deltah_dist}), we first apply a log-transform
followed by
layer-wise z-score normalization:
$\tilde{z}^{(l,j)} = (\log(\DeltaH^{(l,j)} + \epsilon) - \mu_l)
/ \sigma_l$, $\epsilon = 10^{-8}$.
Here $\mu_l$ and $\sigma_l$ denote the mean and standard deviation of
$\log(\DeltaH^{(l,j)} + \epsilon)$ over experts $j$ in layer $l$.
The final score for expert $j$ in layer $l$ is:
\begin{equation}
s(l,j) = (1 - \alpha) \cdot \tilde{z}_{\text{text}}^{(l,j)}
- \alpha \cdot \tilde{z}_{\text{la}}^{(l,j)}
\label{eq:combined_score}
\end{equation}
where $\alpha$ controls the modality balance.
\textbf{Experts with the lowest $s(l,j)$ are pruned first.}

\textbf{Correlation-based $\alpha$ selection.}
Since models exhibit varying degrees of text--vision importance
ranking correlation, the modality balance $\alpha$ is
model-specific; we obtain it from a closed-form expression on calibration statistics.
Let $\bar\rho_{x,y} = \tfrac{1}{L}\sum_l \rho_l(\Delta H_x^{(l)},\,
\Delta H_y^{(l)})$ denote the layer-averaged Spearman correlation
between expert rankings under modalities $x, y \in \{\text{t}, \text{v}, \text{la}\}$,
where $\text{t}$, $\text{v}$, and $\text{la}$ index $\DeltaH$ on text,
full-vision, and low-attention vision token subsets respectively.
We adopt $\alpha = a/(a+b)$ with $a \!:=\! \bar\rho_{\text{t},\text{v}} -
\bar\rho_{\text{t},\text{la}}$ and $b \!:=\! \bar\rho_{\text{v},\text{la}}$ (both
positive on the evaluated models, keeping $\alpha \in [0,1]$;
Table~\ref{tab:per_model_modality_gap}).
Intuitively, $a$ captures the \emph{differential alignment} of
text-side importance with the full-vision versus low-attention
signal: when text and full-vision experts overlap but text and
low-attention experts do not, low-attention vision is complementary
to text and $\alpha$ tends to be larger.
$b$ captures \emph{within-vision redundancy}: when full-vision and
low-attention rankings agree, favoring low-attention adds little
new information and $\alpha$ tends to be smaller.
Section~\ref{sec:ablation_alpha} reports the closed-form $\alpha$
for each model and verifies that these values lie within the empirical safe range; per-model $\bar\rho$ values are reported in
Appendix~\ref{app:modality_gap}.

\textbf{Global pruning.}
Expert redundancy varies across layers and
modalities~\cite{bai2025diep, huang2025modes}, suggesting that a fixed
per-layer budget is suboptimal.
We instead rank all experts across all layers by $s(l,j)$ and prune
the bottom $\lfloor r \cdot N_{\text{total}} \rfloor$ entries, so that
no layer is forced to prune experts that its own score distribution
does not mark as redundant
(Section~\ref{sec:ablation_components}, Appendix~\ref{app:pruning_dist}).
We compare this against uniform per-layer pruning in
Section~\ref{sec:ablation_components}.

Proposition~\ref{prop:kl_bound} justifies $\DeltaH$ only as a
\emph{within-layer} ranking, since $C_l = L_{\Phi_l}^2/4$ differs across
layers and is not estimated. The z-score above serves a separate purpose:
it makes the per-layer score distributions comparable in shape so that
$\DeltaH$ can be used as a global ranking signal without recovering $C_l$
(Section~\ref{sec:ablation_components}, Appendix~\ref{app:pruning_dist}).
The asymmetric text/vision combination
reflects the modality gap of Section~\ref{sec:motivation_proxy}.

\section{Experiments}
\label{sec:experiments}

\subsection{Experimental Setup}
\label{sec:exp_setup}

\textbf{Models and baselines.}
We evaluate on three MoE-based MLLMs:
InternVL3.5-30B-A3B~\cite{wang2025internvl35},
Qwen3-VL-30B-A3B-Instruct~\cite{bai2025qwen3vl}, and
Gemma4-26B-A4B-Instruct~\cite{team2025gemma4}.\footnote{``A$x$B''
denotes ${\sim}x$ billion active parameters per token under
top-$k$ routing (e.g., 30B-A3B = 30B total, 3B active per token).}
All three share the same expert pool for both vision and text tokens;
detailed model configurations are provided in
Appendix~\ref{app:model_config}.
We compare against
(1)~HC-SMoE~\cite{chen2024hcsmoe}, which merges similar
experts via hierarchical clustering, and
(2)~REAP~\cite{lasby2025reap}, which prunes experts by
router-weighted activation norms.

\textbf{Benchmarks.}
We report results on several multimodal benchmarks---ChartQA~\cite{masry2022chartqa},
TextVQA~\cite{singh2019textvqa}, MMStar~\cite{mmstar2024},
MMBench~\cite{liu2023mmbench}, RealWorldQA~\cite{xai2024realworldqa},
MME~\cite{fu2023mme},
POPE~\cite{li2023pope}, and
CV-Bench~\cite{tong2024cambrian} (average of 2D and 3D splits)---
covering chart understanding, visual question answering, hallucination evaluation, and 2D/3D vision-centric
perception.
To measure the effect of pruning on language-only capabilities, we
additionally evaluate on three text-only benchmarks:
MMLU-Redux~\cite{gema2025we}, IFEval~\cite{zhou2023ifeval},
and ARC-Challenge~\cite{clark2018arc}.
Further details on hardware and evaluation pipeline are provided in
Appendix~\ref{app:exp_settings}.

\subsection{Main Results}
\label{sec:main_results}

\begin{table}[t]
\caption{Main results at 25\% and 50\% expert pruning ratios.
$^\dagger$Unpruned-model results are reproduced with our evaluation pipeline.
``Avg. (\%)''\ is the mean retention of each method's per-benchmark
score relative to the unpruned baseline, averaged over all 11 benchmarks.}
\label{tab:main_results}
\centering
\small
\resizebox{\textwidth}{!}{%
\begin{tabular}{@{}l c l@{\hspace{6pt}} cccccccc ccc|c@{}}
\toprule
 & & & \multicolumn{8}{c}{\textit{Multimodal}}
 & \multicolumn{3}{c}{\textit{Text-only}}
 & \multirow{2}{*}{\shortstack{Avg.\\(\%)}} \\
\cmidrule(lr){4-11} \cmidrule(lr){12-14}
Model & Ratio & Method
 & MME & MMBench & MMStar & ChartQA & TextVQA
 & RealWorldQA & CV-Bench & POPE & MMLU-Redux & IFEval & ARC-C
 & \\
\midrule
\multirow{7}{*}{\shortstack{InternVL3.5\\30B-A3B}}
 & --- & Original$^\dagger$
 & 2442.2 & 85.2 & 71.4 & 87.3 & 80.2
 & 72.2 & 84.6 & 89.4 & 80.9 & 78.9 & 95.2 & 100.0 \\
\cmidrule(lr){2-15}
 & \multirow{3}{*}{25\%}
 & HC-SMoE
 & 2344.8 & 79.8 & 67.1 & \textbf{87.0} & 76.2
 & 71.6 & \textbf{84.5} & \textbf{89.8} & 67.4 & 45.6 & 87.4 & 91.9 \\
 & & REAP
 & 2326.6 & \textbf{84.0} & 68.9 & 83.2 & 78.8
 & \textbf{72.5} & 83.6 & 89.3 & 69.3 & 69.7 & 92.7 & 95.9 \\
 & & \method
 & \textbf{2364.9} & 83.2 & \textbf{69.3} & \textbf{87.0}
 & \textbf{78.9} & 71.6
 & 84.3 & 89.7 & \textbf{72.3} & \textbf{77.1} & \textbf{94.1} & \textbf{97.7} \\
\cmidrule(lr){2-15}
 & \multirow{3}{*}{50\%}
 & HC-SMoE
 & 1907.5 & 66.1 & 56.2 & 75.3 & 66.5
 & 68.0 & 79.8 & 88.3 & 33.9 & 20.9 & 53.2 & 74.1 \\
 & & REAP
 & 2109.6 & 78.7 & \textbf{63.3} & 65.3 & 74.4
 & 71.2 & \textbf{83.9} & 89.1 & 53.9 & \textbf{66.4} & 85.7 & 88.5 \\
 & & \method
 & \textbf{2160.3} & \textbf{80.7} & \textbf{63.3} & \textbf{76.7}
 & \textbf{75.6} & \textbf{71.6}
 & 83.3 & \textbf{89.7} & \textbf{59.8} & 64.9 & \textbf{90.6} & \textbf{91.2} \\
\midrule
\multirow{7}{*}{\shortstack{Qwen3-VL\\30B-A3B}}
 & --- & Original$^\dagger$
 & 2416.6 & 87.3 & 66.9 & 85.3 & 84.6
 & 73.5 & 86.0 & 88.5 & 79.0 & 86.0 & 96.7 & 100.0 \\
\cmidrule(lr){2-15}
 & \multirow{3}{*}{25\%}
 & HC-SMoE
 & 2085.4 & 83.9 & 65.3 & \textbf{84.3} & 83.0
 & 71.9 & 81.3 & \textbf{88.8} & \textbf{73.6} & 81.7 & 94.9 & 96.0 \\
 & & REAP
 & 2133.5 & 83.3 & \textbf{65.7} & 84.2 & 82.9
 & 68.4 & 73.9 & 88.3 & 73.3 & 80.9 & 95.1 & 94.8 \\
 & & \method
 & \textbf{2287.3} & \textbf{84.4} & \textbf{65.7} & 82.5 & \textbf{83.9}
 & \textbf{72.5}
 & \textbf{82.8} & 88.4 & 68.7 & \textbf{83.0} & \textbf{95.3} & \textbf{96.6} \\
\cmidrule(lr){2-15}
 & \multirow{3}{*}{50\%}
 & HC-SMoE
 & 1923.9 & 75.1 & 59.9 & \textbf{74.3} & 76.1
 & 69.4 & 80.5 & 85.2 & \textbf{56.0} & 65.2 & 74.8 & 85.5 \\
 & & REAP
 & 2026.6 & \textbf{80.3} & 59.9 & 67.1 & 78.1
 & 66.0 & 70.1 & 88.2 & \textbf{56.0} & \textbf{73.8} & \textbf{85.9} & 86.6 \\
 & & \method
 & \textbf{2042.8} & \textbf{80.3} & \textbf{60.2} & 69.6
 & \textbf{79.7} & \textbf{69.7}
 & \textbf{85.2} & \textbf{89.5} & 53.0 & 69.3 & 83.1 & \textbf{88.3} \\
\midrule
\multirow{7}{*}{\shortstack{Gemma4\\26B-A4B}}
 & --- & Original$^\dagger$
 & 2333.2 & 88.9 & 71.1 & 69.2 & 71.8
 & 76.3 & 85.1 & 88.1 & 80.3 & 88.2 & 97.1 & 100.0 \\
\cmidrule(lr){2-15}
 & \multirow{3}{*}{25\%}
 & HC-SMoE
 & 1970.1 & 22.0 & 37.4 & 66.7 & 63.8
 & 53.1 & 71.3 & 85.5 & 62.2 & 67.3 & 94.9 & 77.2 \\
 & & REAP
 & 2204.9 & 84.9 & \textbf{65.8} & 66.6 & \textbf{72.1}
 & 75.1 & 82.6 & \textbf{88.3} & \textbf{70.6} & 83.7 & \textbf{95.9} & \textbf{96.0} \\
 & & \method
 & \textbf{2224.9} & \textbf{85.8} & 64.6 & \textbf{68.4} & 70.8
 & \textbf{76.2}
 & \textbf{84.0} & 88.2 & 64.5 & \textbf{86.5} & 95.5 & \textbf{96.0} \\
\cmidrule(lr){2-15}
 & \multirow{3}{*}{50\%}
 & HC-SMoE
 & 553.4 & 9.8 & 27.6 & 35.6 & 14.0
 & 40.1 & 49.0 & 49.6 & 25.5 & 26.1 & 58.7 & 39.3 \\
 & & REAP
 & 1939.3 & 61.0 & 50.7 & \textbf{63.4} & 65.2
 & 66.8 & 75.9 & 87.9 & 41.0 & 67.8 & 77.7 & 80.9 \\
 & & \method
 & \textbf{1975.4} & \textbf{78.2} & \textbf{52.3} & 54.4 & \textbf{70.7}
 & \textbf{72.3} & \textbf{82.7} & \textbf{88.4} & \textbf{51.8} & \textbf{73.6} & \textbf{88.1} & \textbf{86.7} \\
\bottomrule
\end{tabular}%
}
\end{table}

As shown in Table~\ref{tab:main_results}, \method retains
near-original performance at 25\% pruning and achieves the best or tied-best average retention on every model--ratio configuration, with the gap widening as the pruning
ratio increases.

\textbf{Multimodal benchmarks.}
At 25\% pruning, \method stays close to the unpruned baseline across
models, with the best or tied-best average retention on every configuration
(``Avg.''\ column in Table~\ref{tab:main_results}).
At 50\%, \method maintains the highest average retention across all
three models, while HC-SMoE drops sharply and REAP trails consistently.
The advantage of modality-aware pruning grows with sparsity: at
$25\%$ REAP stays within ${\sim}2$ points of \method, while at
$50\%$ the gap widens to $2$--$6$ points, reflecting that the cost
of modality-uniform treatment grows with sparsity.
HC-SMoE's similarity-based merging is competitive at 25\% on
InternVL3.5 and Qwen3-VL but drops sharply on Gemma4, where its
modality-uniform clustering conflates experts with distinct
modality roles.
REAP's router-weighted activation scoring degrades more gradually
but treats vision and text tokens uniformly, missing the asymmetric
importance documented in Section~\ref{sec:motivation} and yielding
consistently lower multimodal retention than \method.
\method's text/vision decomposition addresses both: text-side
preservation protects language-critical experts, while
low-attention vision scoring isolates vision-side redundancy.

\textbf{Text-only benchmarks.}
While all methods suffer substantial degradation on text-only benchmarks at 50\% pruning, 
\method remains competitive on most cells and achieves the highest aggregate retention 
despite trailing REAP on the three Qwen3-VL text-only benchmarks
(separate multimodal and text-only averages are given in Appendix~\ref{app:separate_averages}).
\subsection{Ablation Study and Analysis}
\label{sec:ablation}

\subsubsection{Modality Decomposition, Vision Signal, and Pruning Strategy}
\label{sec:ablation_components}

\textbf{Modality decomposition.}
Table~\ref{tab:ablation}(a) shows that the modality
decomposition itself is essential: aggregating $\DeltaH$
uniformly over all tokens (no text/vision split) substantially
degrades MMStar, ChartQA, and TextVQA, indicating that the
asymmetric text-preserve / vision-prune signaling cannot be
recovered from a single undifferentiated importance score.

\textbf{Vision signal alternatives.}
Table~\ref{tab:ablation}(a) confirms that low-attention's
dataset-stability advantage over the high-attention and
full-vision alternatives is reflected in downstream performance:
replacing it with either alternative substantially degrades visual reasoning benchmarks.
For these alternatives we flip the sign in
Eq.~\ref{eq:combined_score} (\,$-\alpha\,\tilde{z}_{\text{la}}
\rightarrow +\alpha\,\tilde{z}_{\text{vision}/\text{ha}}$\,)
because high-attention and full-vision signals capture
\emph{important} vision tokens (preservation signal), whereas
low-attention captures \emph{unimportant} ones (pruning signal).
Overall, high-attention is comparable on
benchmarks close to the calibration distribution (MME, TextVQA)
but drops substantially on distant benchmarks like ChartQA ($-22.2$).
Appendix~\ref{app:vision_selectors} extends this comparison to
representation-drift and diversity-based token selectors.

\begin{figure}[t]
\centering
\includegraphics[width=0.85\textwidth]{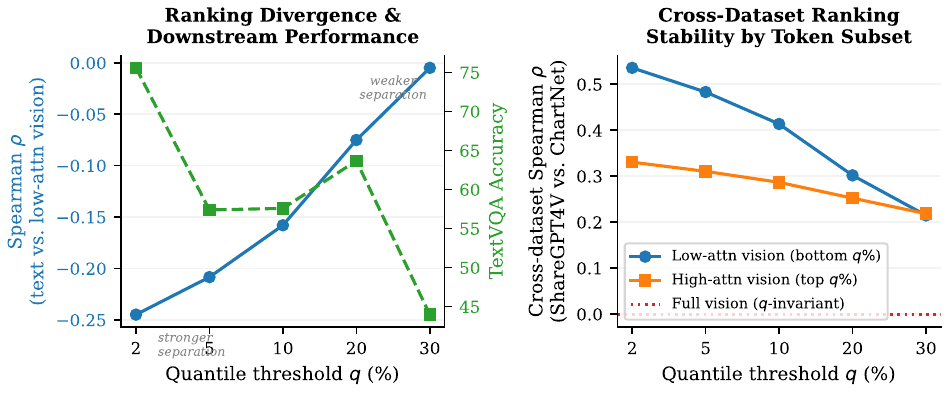}
\caption{Effect of quantile threshold $q$ on InternVL3.5 at 50\%
pruning.
Left: Spearman $\rho$ between text and low-attention vision rankings
(more negative = more distinct) and TextVQA accuracy.
Right: cross-dataset Spearman $\rho$ (ShareGPT4V vs.\ ChartNet) by
token subset under matched-size symmetric quantiles (low-attn =
bottom $q\%$, high-attn = top $q\%$).}
\label{fig:quantile_ablation}
\end{figure}

\textbf{Low-attention threshold.}
We vary the quantile threshold $q$ used to define low-attention vision
tokens, considering $q \in \{2\%, 5\%, 10\%, 20\%, 30\%\}$.
As $q$ increases, more semantically relevant tokens enter the
low-attention set, making the vision-side ranking converge toward
the text-side ranking.
Figure~\ref{fig:quantile_ablation} (left) confirms this: $\rho$ rises
from negative correlation at $q{=}2\%$ to near zero at $q{=}30\%$,
with TextVQA accuracy tending to decrease as $q$ grows---modality-decomposed scoring (Eq.~\ref{eq:combined_score}) relies on
the two signals carrying distinct information.
The right panel further shows that low-attention vision tokens yield
the most dataset-stable expert rankings: cross-dataset $\rho$
(ShareGPT4V vs.\ ChartNet) reaches $0.54$ at $q{=}2\%$, compared to
$0.33$ for the size-matched high-attention subset (top $2\%$) and
${\approx}0$ for full vision tokens.
This suggests that the strictest threshold isolates vision tokens
whose expert importance is both modality-discriminative and
dataset-stable.

\begin{table}[t]
\centering
\footnotesize
\caption{Ablation studies.
(a)~Scoring components and pruning strategy on InternVL3.5
(50\% pruned).
(b)~Calibration set design on Qwen3-VL (50\% pruned): composition and
number of samples.
Bold marks the best value per column; \emph{collapse} indicates
average retention below $30\%$ of the unpruned baseline.
$^{\dagger}$$\alpha$ re-tuned for the uniform budget
(Section~\ref{sec:ablation_components}).}
\label{tab:ablation}
\begin{minipage}[t]{0.51\textwidth}
\centering
\scriptsize
\textbf{(a)} Scoring components and pruning strategy\\
(InternVL3.5, 50\% pruned)\\[0.3em]
\setlength{\tabcolsep}{3pt}
\resizebox{\linewidth}{!}{%
\begin{tabular}{@{}l cccc@{}}
\toprule
Configuration & MME & MMStar & ChartQA & TextVQA \\
\midrule
Default                        & 2160 & \textbf{63.3} & \textbf{76.7} & 75.6 \\
\midrule
\multicolumn{5}{@{}l}{\emph{Modality decomposition}} \\
w/o modality split             & 2034 & 51.1 & 41.2 & 66.0 \\
\midrule
\multicolumn{5}{@{}l}{\emph{Vision-side signal}} \\
w/ high-attn (top $\{2,5,10\}\%$) & \multicolumn{4}{c}{\textcolor{gray}{\emph{collapse}}} \\
w/ high-attn (top $20\%$)         & \textbf{2226} & 55.5 & 54.5 & 76.1 \\
w/ full vision                    & 2152 & 48.3 & 38.8 & 66.9 \\
\midrule
\multicolumn{5}{@{}l}{\emph{Pruning strategy}} \\
w/ uniform per-layer budget ($\alpha{=}0.5$)$^{\dagger}$ & 2104 & 62.2 & 72.1 & \textbf{76.5} \\
\bottomrule
\end{tabular}}
\end{minipage}%
\hfill
\begin{minipage}[t]{0.46\textwidth}
\centering
\scriptsize
\textbf{(b)} Calibration set design\\
(Qwen3-VL, 50\% pruned)\\[0.3em]
\setlength{\tabcolsep}{3pt}
\resizebox{\linewidth}{!}{%
\begin{tabular}{@{}l cccc@{}}
\toprule
Calibration setting & MME & MMStar & ChartQA & TextVQA \\
\midrule
Default (ShareGPT4V, 1K)     & 2042.8 & 60.2 & 69.6 & 79.7 \\
\midrule
\multicolumn{5}{@{}l}{\emph{Varying composition (1K)}} \\
+ ChartNet (7:3)             & 2034.2 & \textbf{64.3} & \textbf{83.5} & 79.4 \\
\midrule
\multicolumn{5}{@{}l}{\emph{Varying \# samples (ShareGPT4V)}} \\
100                          & 2006.0 & 58.7 & 68.2 & 78.9 \\
500                          & \textbf{2047.2} & 60.7 & 66.1 & \textbf{80.3} \\
\bottomrule
\end{tabular}}
\end{minipage}
\end{table}

\textbf{Global vs.\ uniform pruning.}
Prior work has shown that pruning sensitivity is highly non-uniform
across layers~\cite{huang2025layerwisesparsity, zhang2024finercut},
and expert redundancy likewise varies substantially across MoE
layers~\cite{bai2025diep}.
Uniform pruning, which removes the same number of experts from
every layer, ignores this non-uniformity; in contrast, our global
ranking exploits per-layer differences in score-distribution shape
(layer-wise z-scoring centers each layer's $\tilde z$ at zero, so
the cross-layer pruning rate reflects skewness and tail thickness,
not absolute magnitude).
Table~\ref{tab:ablation}(a) shows that global allocation still leads
uniform pruning on MME and ChartQA on InternVL3.5, even with $\alpha$
re-tuned for the uniform budget ($\alpha{=}0.5$; a forced per-layer
quota leaves more of the protection to the text term, so it calls for
a smaller $\alpha$; full sweep in Appendix~\ref{app:scoring_allocation}).
Appendix~\ref{app:pruning_dist} visualizes the resulting per-layer
distribution.

\subsubsection{Calibration Set Design: Domain and Size}
\label{sec:ablation_calibration}

With the default ShareGPT4V calibration, Qwen3-VL and Gemma4 at 50\% pruning
show a pronounced weakness on ChartQA.
We trace this to \textit{calibration domain mismatch}: ShareGPT4V
contains primarily natural-image conversations with little chart or
document content, so experts specialized for chart reasoning receive
limited activation during calibration and are consequently under-scored
by $\DeltaH$.
This implies that \method's importance scores are responsive to
deployment-domain signal in the calibration set---a property
practitioners can exploit to target a specific deployment
distribution.

To demonstrate this, we construct a mixed calibration set of 700
ShareGPT4V~\cite{chen2023sharegpt4v} and 300
ChartNet~\cite{ibm2024chartnet} reasoning samples, where ChartNet
pairs chart images with extended reasoning text that matches
ShareGPT4V's text-token density.
Under this mixed calibration, \method on Qwen3-VL at 50\% pruning
(Table~\ref{tab:ablation}(b)) recovers \textbf{13.9 points on ChartQA}, while also improving MMStar by 4.1 points;
MME and TextVQA are nearly identical.
The absence of a substantial trade-off supports two claims:
(i)~the ChartQA gap is attributable to calibration domain mismatch; and
(ii)~\method can be \emph{targeted} at a deployment domain by
augmenting calibration with in-domain data.

This is consistent with the REAP's~\cite{lasby2025reap} strategy of combining
general pre-training corpora with domain-specific data for
calibration: a diverse mixture prevents catastrophic drops on
specialized distributions and, as we observe, can further improve
in-domain performance. We leave to future work calibration-mixing
strategies and extension to reasoning-heavy generative
benchmarks (e.g., coding, mathematical problem solving).

\textbf{Calibration size.}
Beyond composition, we test how far the calibration budget can be
reduced by subsampling ShareGPT4V from the default 1{,}000 down to 100
samples (Table~\ref{tab:ablation}(b), bottom block).
These results suggest \method is robust to calibration set
size: even at 100 samples, \method retains most of the default-1K
performance, with some fluctuation on individual benchmarks (e.g.,
ChartQA).

\subsubsection{\texorpdfstring{$\alpha$}{alpha} Sensitivity and Safe Range}
\label{sec:ablation_alpha}

Section~\ref{sec:method} fixes $\alpha$ deterministically, which yields $0.75 / 0.33 / 0.08$ for InternVL3.5 / Qwen3-VL / Gemma4---rounded to $0.8 / 0.3 / 0.1$ (the ``Used $\alpha$'' column in Table~\ref{tab:alpha_sens}) for simplicity. 
We verify that these values lie within an empirically stable range by sweeping $\alpha$ at both pruning ratios and reporting the average performance retention over MME, MMStar, ChartQA, and TextVQA.
Table~\ref{tab:alpha_sens} shows that the empirical safe range
narrows for models with higher within-vision redundancy
(the $b := \bar\rho_{\text{v},\text{la}}$ feature in Section~\ref{sec:method}):
InternVL3.5 tolerates a wide range of $\alpha$, Qwen3-VL exhibits
a sharp degradation once $\alpha$ exceeds its operating point, and
Gemma4 degrades gradually as $\alpha$ grows.

\begin{wraptable}{r}{0.55\textwidth}
\vspace{-1.0em}
\centering
\scriptsize
\caption{$\alpha$ sensitivity at 25/50\% pruning.
Cells: \% retention of unpruned baseline
(avg.\ over MME, MMStar, ChartQA, TextVQA).}
\label{tab:alpha_sens}
\setlength{\tabcolsep}{2pt}
\begin{tabular}{@{}l c c ccccc}
\toprule
Model & Used~$\alpha$ & Ratio & \multicolumn{5}{c}{$\alpha$ sweep (\% retention)} \\
\cmidrule(lr){4-8}
      &              &        & 0.1 & 0.3 & 0.5 & 0.7 & 0.9 \\
\midrule
\multirow{2}{*}{InternVL3.5} & \multirow{2}{*}{0.8}  & 25\% & 97.5 & 98.3 & 98.1 & 98.0 & 98.1 \\
                             &                       & 50\% & 84.3 & 85.9 & 86.9 & 89.1 & 65.1 \\
\midrule
\multirow{2}{*}{Qwen3-VL}    & \multirow{2}{*}{0.3}  & 25\% & 96.4 & 97.1 & 97.1 & 29.0 & 26.0 \\
                             &                       & 50\% & 84.4 & 87.6 & 25.4 & 29.6 & 23.0 \\
\midrule
\multirow{2}{*}{Gemma4}      & \multirow{2}{*}{0.1}  & 25\% & 95.9 & 91.1 & 72.5 & 67.6 & 66.7 \\
                             &                       & 50\% & 83.8 & 74.1 & 60.9 & 51.1 & 41.0 \\
\bottomrule
\end{tabular}
\vspace{-0.8em}
\end{wraptable}
This is consistent with the redundancy interpretation of $b$:
when full-vision and low-attention rankings agree, a large $\alpha$
overemphasizes a vision-side signal that offers little additional
information.
Across all three models, the closed-form $\alpha$ lies within a
stable region of the sensitivity sweep: it coincides with the
row's top-performing swept value for Qwen3-VL and Gemma4, and for
InternVL3.5 it lies at the upper end of the stable region
($\alpha\le0.7$ in the sweep; retention collapses at $\alpha{=}0.9$),
consistent with the main results in Table~\ref{tab:main_results}.
We adopt a single fixed $\alpha$ per model for simplicity; exploring
layer-wise modality weights is a promising direction, since modality
sensitivity may vary across layers.

\section{Conclusion}
\label{sec:conclusion}

We presented \method, a training-free framework for pruning MoE
experts in MLLMs by decomposing their importance into text and vision
contributions.
By combining a redistribution-aware importance metric ($\DeltaH$) with
low-attention vision tokens as a stable pruning signal, \method
helps identify which experts can be removed with reduced risk of
disproportionately degrading either modality.
Experiments across multiple models and benchmarks show that this
modality-aware approach retains near-original performance at 25\%
pruning and generally outperforms modality-agnostic baselines at
50\% pruning, reducing the expert-weight footprint while
preserving stronger performance than competing compression methods. This work inspires further research on modality-aware compression strategies
for MoE-based MLLMs and enables deployment in
resource-constrained settings.

\begin{ack}
The author thanks KyungHwan Shim, Huiseong Gim, and JinHyeok Park of the NetsPresso
team for their help and valuable discussions.
This work was supported by Nota Inc. and by the Technology Innovation Program
(RS-2024-00468747, Development of AI and Lightweight Technology for
Embedding Multisensory Intelligence Modules) funded by the Ministry of
Trade Industry \& Energy (MOTIE, Korea).
\end{ack}

\bibliographystyle{plainnat}
\bibliography{references}

\newpage
\appendix

\section{Limitations}
\label{app:limitations}

\method applies a fixed prune mask per layer, without modeling the
joint removal of experts that back each other up or input-conditional
sparsity. Moreover, the adaptive
$\alpha$ rule is empirically validated rather than derived from a
theoretical optimum. Finally, \method also targets the training-free
setting; fine-tuning (e.g., LoRA~\cite{hu2022lora}) is orthogonal and
could recover residual gaps.

\textbf{Pairwise redundancy and one-shot scoring.}
$\DeltaH$ is not a fully independent score: its redistribution
counterfactual (Section~\ref{sec:deltah}) measures the
residual output change after the remaining co-activated experts absorb
the removed expert's routing mass, so first-order redundancy among
co-activated experts is captured implicitly. What it does not model is
the joint removal of experts that back each other up. To quantify how
much this matters at our operating ratio, we compared the one-shot
50\% prune mask on InternVL3.5 against a two-stage variant that prunes
25\%, re-scores $\DeltaH$ on the pruned model, and then prunes to 50\%
in total. The two masks differ substantially (Jaccard $0.77$; 800 of
the 3{,}472 experts in their union differ), but the two-stage variant
is strictly worse (one-shot $\rightarrow$ two-stage:
MME $2160.3\rightarrow1956.7$, MMStar $63.3\rightarrow53.9$, ChartQA
$76.7\rightarrow54.3$, TextVQA $75.6\rightarrow64.9$). We attribute
this to re-scoring on an already-redistributed model: $\DeltaH$
re-measured after partial pruning no longer reflects the original
model's redundancy structure, whereas the one-shot score models that
redistribution analytically without altering the model it measures.
This experiment therefore does not isolate interaction effects, and
we leave them, as well as input-adaptive or iterative pruning at more
aggressive ratios, open.

\textbf{Calibration-domain sensitivity.}
As with other calibration-based post-training compression methods,
\method's prune decisions depend on the calibration distribution:
domains that the calibration set under-represents can lead to
under-scored experts (Section~\ref{sec:ablation_calibration}). This
sensitivity is confined to the text-side preservation signal, whereas
the low-attention vision signal is stable across calibration datasets
(Appendix~\ref{app:calib_swap}); it is therefore controllable rather
than intrinsic, and we do not claim domain invariance. We recommend
(i)~a diverse calibration mixture as the default; (ii)~augmenting it
with in-domain data when the deployment domain is known
(Table~\ref{tab:ablation}(b)); and (iii)~when the domain is unknown,
broadening the mixture with chart-, document-, and OCR-style data as an
inexpensive safeguard against under-coverage.

Promising future directions include layer-wise $\alpha$ adaptation,
calibration-domain mixing strategies for deployment specialization,
and stacking with orthogonal compression families
(quantization, intra-expert decomposition).

\section{Proof of the Motivational Bound (Proposition~\ref{prop:kl_bound})}
\label{app:proof}

\begin{proposition*}
Let $\Phi_l:\mathbb{R}^{T\times d}\to\mathbb{R}^{T\times |V|}$ map the
layer-$l$ MoE outputs of a $T$-token sequence to the pre-softmax
logits, with Lipschitz constant $L_{\Phi_l}$ (Frobenius norm) on the
activation domain induced by RMSNorm/LayerNorm.  With
$\hat h_t := \HMoE^{(l)}(h_t)$ and
$\hat h_t^{(-j)} := \HMoE^{(l,-j)}(h_t)$ the MoE outputs at position
$t$ before/after removing expert $j$,
$\hat H := (\hat h_1,\dots,\hat h_T)$ and $\hat H^{(-j)}$ the
corresponding sequences, and
$p_t := \mathrm{softmax}(\Phi_l(\hat H)_t)$,
$p_t^{(-j)} := \mathrm{softmax}(\Phi_l(\hat H^{(-j)})_t)$,
\[
\frac{1}{T}\sum_{t=1}^{T}
D_{\mathrm{KL}}\!\bigl(p_t\,\|\,p_t^{(-j)}\bigr)
\leq \tfrac{L_{\Phi_l}^2}{4}\,\frac{1}{T}\sum_{t=1}^{T}
\|\hat h_t-\hat h_t^{(-j)}\|_2^2 .
\]
\end{proposition*}

\begin{proof}
Set $\Delta\hat H := \hat H-\hat H^{(-j)}\in\mathbb{R}^{T\times d}$,
whose rows are $\hat h_t-\hat h_t^{(-j)}$, and
$\Delta Z := \Phi_l(\hat H)-\Phi_l(\hat H^{(-j)})\in\mathbb{R}^{T\times|V|}$,
whose rows $\Delta z_t$ are the logit perturbations at position $t$.

\paragraph{Step 1 (logit perturbation).}
Self-attention couples tokens within a sequence, so we view
$\Phi_l$ as a sequence-level map
$\mathbb{R}^{T \times d}\!\to\!\mathbb{R}^{T \times |V|}$ and bound
$\|\Delta Z\|_F$ in terms of $\|\Delta\hat H\|_F$.
Each subsequent residual block $f_i(X)=X+g_i(X)$ is
$(1+L_{g_i})$-Lipschitz in the Frobenius norm on the bounded
activation domain, where $L_{g_i}$ is the local Lipschitz constant of
the attention+FFN branch $g_i$ as a map on the whole sequence; the
final RMSNorm/LayerNorm before the LM head acts row-wise and is
locally $L_{\mathrm{norm}}$-Lipschitz there, and the LM head acts
row-wise with constant $\|W\|_{\mathrm{op}}$ (for row-wise maps the
per-row $\ell_2$ Lipschitz constant is also the Frobenius constant);
composing,
\begin{equation}
\|\Delta Z\|_F
\;\leq\; \|W\|_{\mathrm{op}} \cdot L_{\mathrm{norm}}
  \!\prod_{i=l+1}^{L}\!(1+L_{g_i})\,\|\Delta\hat H\|_F
\;=:\; L_{\Phi_l}\,\|\Delta\hat H\|_F,
\label{eq:perturb}
\end{equation}
where $\|\Delta Z\|_F^2=\sum_t\|\Delta z_t\|_2^2$ and
$\|\Delta\hat H\|_F^2=\sum_t\|\hat h_t-\hat h_t^{(-j)}\|_2^2$.

\paragraph{Step 2 (logit-to-KL, per position).}
Let $A(z):=\log\sum_i e^{z_i}$, $p:=\mathrm{softmax}(z)$,
$q:=\mathrm{softmax}(z')$. Then $\nabla A(z)=p$ and
$\nabla^2 A(z)=\mathrm{diag}(p)-pp^\top$.
For any unit $v$, Popoviciu's inequality gives
$\mathrm{Var}_p[v]\leq\tfrac14(\max v-\min v)^2$, and
$(\max v-\min v)^2\leq 2(a^2+b^2)\leq 2\|v\|_2^2=2$
($a:=\max v$, $b:=\min v$; the first step uses $(a-b)^2\le 2(a^2+b^2)$
and the second that $a,b$ are two coordinates of the unit vector $v$),
so $A$ is $\tfrac12$-smooth.
By Bregman/KL duality~\cite{banerjee2005clustering} and the descent
lemma,
\begin{equation}
D_{\mathrm{KL}}(p\|q)
= A(z')-A(z)-\nabla A(z)^\top(z'-z)
\leq \tfrac{1}{4}\|\Delta z\|_2^2.
\label{eq:kl_logit}
\end{equation}
Applied at position $t$ with $z=\Phi_l(\hat H)_t$ and
$z'=\Phi_l(\hat H^{(-j)})_t$, this gives
$D_{\mathrm{KL}}(p_t\|p_t^{(-j)})\leq\tfrac14\|\Delta z_t\|_2^2$ for
every $t$.

\paragraph{Step 3 (combine).}
Summing the per-position bound of Step 2 over the $T$ positions and
applying~\eqref{eq:perturb},
\[
\sum_{t=1}^{T} D_{\mathrm{KL}}\bigl(p_t\,\|\,p_t^{(-j)}\bigr)
\;\leq\; \tfrac14\|\Delta Z\|_F^2
\;\leq\; \tfrac{L_{\Phi_l}^2}{4}\,\|\Delta\hat H\|_F^2
\;=\; \tfrac{L_{\Phi_l}^2}{4}\sum_{t=1}^{T}\|\hat h_t-\hat h_t^{(-j)}\|_2^2 ,
\]
which, divided by $T$, is the proposition.
Averaging over the calibration distribution (samples
$x\sim\mathcal{D}$ with $|x|$ tokens $h \in x$), for each expert $j$,
\begin{equation}
\mathbb{E}_{x \sim \mathcal{D}}\!\Big[\tfrac{1}{|x|}\!\sum_{h \in x}
  D_{\mathrm{KL}}\Big]
\;\leq\; \tfrac{L_{\Phi_l}^2}{4}\,
  \mathbb{E}_{x \sim \mathcal{D}}\!\Big[\tfrac{1}{|x|}\!\sum_{h \in x}
  \|\hat h-\hat h^{(-j)}\|_2^2\Big]
\;=\; \tfrac{L_{\Phi_l}^2}{4}\,\DeltaH^{(l,j)},
\end{equation}
where $L_{\Phi_l}:=\sup_x L_{\Phi_l(\,\cdot\,;x)}<\infty$ on the
bounded activation domain.
\end{proof}

\textbf{Remark (layer-depth effect).}
$L_{\Phi_l}$ contains $\prod_{i=l+1}^{L}(1+L_{g_i})$, which contracts
as $l\to L$ (with $L_{\Phi_L}=\|W\|_{\mathrm{op}}L_{\mathrm{norm}}$); the bound
tightens for deeper layers and the spectral-norm bound
$\lambda_{\max}(\mathrm{diag}(p)-pp^\top)\leq 1/2$ is loose for
sharp vocabulary distributions, both consistent with the empirical
findings of Section~\ref{sec:kl_bound} and Appendix~\ref{app:spearman_depth}.

\section{Model Configurations and Memory Analysis}
\label{app:model_config}

\begin{table}[!ht]
\centering
\caption{MoE model configurations and memory footprint under expert
pruning (BF16).}
\label{tab:model_config}
\small
\setlength{\tabcolsep}{3pt}
\renewcommand{\arraystretch}{0.95}
\begin{tabular}{lccccccccc}
\toprule
\multicolumn{7}{c}{Configuration}
& \multicolumn{3}{c}{Memory (GB)} \\
\cmidrule(lr){1-7} \cmidrule(lr){8-10}
Model & Total & Expert & Layers & E/L & Top-$k$ & Router
& BF16 & 25\% & 50\% \\
\midrule
InternVL3.5
& 30.8B & 29.0B (94\%) & 48 & 128 & 8 & softmax
& 57.5 & 44.0 (-23\%) & 30.5 (-47\%) \\
Qwen3-VL
& 31.1B & 29.0B (93\%) & 48 & 128 & 8 & softmax
& 57.9 & 44.4 (-23\%) & 30.9 (-47\%) \\
Gemma4-26B-A4B
& 25.8B & 22.8B (88\%) & 30 & 128 & 8 & softmax
& 48.1 & 37.4 (-22\%) & 26.8 (-44\%) \\
\bottomrule
\end{tabular}
\end{table}

In Table~\ref{tab:model_config}, memory columns are reported in GB (BF16 weights).
Expert parameters constitute the dominant share of total model size
across all three models.
For InternVL3.5 and Qwen3-VL, 50\% pruning reduces the BF16 weight
footprint from ${\sim}$58\,GB to ${\sim}$31\,GB, placing model weights
within the nominal capacity of a single 40\,GB GPU.
Actual deployment capacity additionally depends on KV-cache size, the
vision tower and projector, activations, framework overhead, context
length, and batch size.
Gemma4-26B-A4B shares a similar MoE configuration (128 experts,
top-8) with a shallower backbone (30 layers), providing architectural
diversity for validating \method across different model scales.

\subsection{Measured Deployment Metrics}
\label{app:deployment}

Table~\ref{tab:deployment} reports end-to-end measurements for
InternVL3.5-30B-A3B on a single NVIDIA A100 80GB (HuggingFace
Transformers with \texttt{accelerate}; greedy decoding).
Since top-$k$ is unchanged by pruning, per-token FLOPs and
single-sequence latency are invariant by design, and we confirm this
empirically: the $B{=}1$ decode-throughput ratio (50\%-pruned /
unpruned) is at parity within run-to-run noise.
The memory saving translates into throughput in two serving regimes.
(a)~\emph{Memory-capped serving}: under a 40\,GB device-map cap, the
unpruned model must offload ${\sim}33\%$ of its parameters to CPU,
whereas the 50\%-pruned model fits fully resident and decodes
$3.9\times$ faster.
(b)~\emph{Large-batch serving}: MoE decode is bandwidth-bound, and the
per-step cost tracks the number of unique activated experts, which
grows with the batch size and is bounded by the size of the expert
pool; halving the pool lowers this ceiling, and the decode-throughput
ratio rises from parity at $B{=}1$ to $1.37\times$ at $B{=}64$ with
diverse-prompt batches (identical-prompt batches keep the per-step
expert union at top-$k$ and show parity, consistent with the
mechanism).
These measurements use HuggingFace's non-fused MoE kernels; fused
grouped-GEMM engines are more strictly bandwidth-bound and would likely
amplify effect~(b).
Loading time is dominated by weight I/O and shrinks with the serialized
checkpoint, which is 27\,GB smaller at 50\% pruning.

\begin{table}[ht]
\centering
\small
\caption{Measured deployment metrics for InternVL3.5-30B-A3B (A100
80GB; HuggingFace Transformers + \texttt{accelerate}).
Protocol: greedy decoding of a fixed short prompt (${\sim}25$ tokens
after the chat template), generating 128 new tokens when resident and
32 when offloaded (tok/s is rate-normalized; the wall clock includes the
single prefill pass); the batch sweep generates 64 new tokens per
sequence; warm-up runs excluded, CUDA-synchronized timing, averaged
over 2--3 runs.}
\label{tab:deployment}
\resizebox{\textwidth}{!}{%
\begin{tabular}{@{}ll@{}}
\toprule
Metric & Measured \\
\midrule
Resident weight memory at 0 / 25\% / 50\% pruning & 57.5 / 44.0 / 30.5\,GB ($-47\%$ at 50\%) \\
$B{=}1$ decode-throughput ratio (50\%-pruned / unpruned) & 0.96--1.02 (parity within run-to-run noise) \\
Decode under a 40\,GB memory cap & unpruned (${\sim}33\%$ CPU-offloaded): 1.52\,tok/s; 50\%-pruned (fully resident): 5.88\,tok/s ($3.9\times$) \\
Large-batch decode-throughput ratio (50\%-pruned / unpruned) & 1.13 ($B{=}8$) / 1.29 ($B{=}16$, $B{=}32$) / 1.37 ($B{=}64$) \\
Unique activated experts per layer per decode step at $B{=}64$ & ${\sim}55$ (unpruned) vs.\ ${\sim}39$ (50\%-pruned) \\
\bottomrule
\end{tabular}}
\end{table}

\section{Experimental Settings}
\label{app:exp_settings}

\subsection{Implementation Details}

\textbf{Hardware.}
All experiments are conducted on a single NVIDIA A100 80GB GPU.
Calibration requires a single forward pass over the 1K calibration
samples per model (no back-propagation or optimizer state); the
remaining compute is benchmark inference.

\textbf{Calibration Data.}
Following~\cite{li2025mbq}, which shows that image captioning data
provides balanced coverage of both visual and linguistic processing,
we construct the calibration set from
ShareGPT4V~\cite{chen2023sharegpt4v}.
Samples are sorted in descending order of text length; the top 1K
samples form the \textbf{calibration set} used for modality-decomposed importance scoring.
The set does not overlap with the downstream evaluation benchmarks.
Longer text sequences provide richer token-level statistics for expert importance estimation, as they activate a broader range of expert routing patterns during the prefill stage.
We report cross-calibration-source variants in Section~\ref{sec:ablation_calibration}.

\textbf{HC-SMoE Configuration.}
We use HC-SMoE with hierarchical agglomerative clustering on
expert-output similarity, average linkage, and frequency-weighted
merging
(\texttt{similarity-base}=\texttt{expert-output},
\texttt{cluster-method}=\texttt{hierarchical},
\texttt{linkage}=\texttt{average},
\texttt{merge-method}=\texttt{freq}).
This corresponds to the default configuration in
\cite{chen2024hcsmoe}, applied to MoE-MLLM backbones in our setting.

\textbf{REAP Configuration.}
We follow the original REAP~\cite{lasby2025reap} formulation,
sharing the same calibration set as \method, and apply
uniform per-layer pruning at the target ratio.

\textbf{DiEP Configuration (Appendix~\ref{app:diep}).}
DiEP~\cite{bai2025diep} is reimplemented in our pipeline; on
synthetic MoE blocks, the reimplementation reproduces the reference
code's MoE forward outputs and router logits bit-exactly and selects
the same prune map for a given budget.
We use DiEP's published hyperparameters (128 calibration sequences
drawn from the same ShareGPT4V calibration set as \method, 10 epochs,
learning rate $5{\times}10^{-3}$ with a cosine schedule, sparsity
weight $\lambda{=}0.01$) and the identical evaluation pipeline and
judge.
DiEP's differentiable search is ratio-independent, so a single search
yields both the 25\% and 50\% masks; every layer keeps at least
top-$k$ experts.
Text-only benchmarks were not run for DiEP.

\subsection{Evaluation Details}

We evaluate on 11 benchmarks spanning two categories:

\textbf{Multimodal Benchmarks.}
\begin{itemize}
    \item \textbf{ChartQA}~\cite{masry2022chartqa}:
    Chart understanding with extractive and reasoning questions.
    Metric: relaxed accuracy.

    \item \textbf{TextVQA}~\cite{singh2019textvqa}:
    Visual question answering requiring OCR of scene text.
    Metric: VQA accuracy.

    \item \textbf{MMStar}~\cite{mmstar2024}:
    Hand-curated, vision-indispensable multi-discipline reasoning samples with minimal data leakage.
    Metric: accuracy.

    \item \textbf{MMBench}~\cite{liu2023mmbench}:
    Comprehensive multimodal evaluation across 20 ability dimensions. We use the \textsc{MMBench\_DEV\_EN\_V11} split (English dev set,
    version 1.1) 
    Metric: accuracy.

    \item \textbf{RealWorldQA}~\cite{xai2024realworldqa}:
    Real-world spatial understanding and reasoning from natural images.
    Metric: accuracy.

    \item \textbf{MME}~\cite{fu2023mme}:
    Perception and cognition evaluation via yes/no questions.
    Metric: perception + reasoning score.

    \item \textbf{CV-Bench}~\cite{tong2024cambrian}:
    Vision-centric perception benchmark with 2D (object counting,
    relation, spatial relations) and 3D (depth, distance) splits;
    we report the average of the 2D and 3D scores.
    Metric: accuracy.

    \item \textbf{POPE}~\cite{li2023pope}:
    Object hallucination evaluation via polling-based binary questions.
    Metric: F1 score.
\end{itemize}

\textbf{Text-Only Benchmarks.}
We additionally include text-only benchmarks to measure the effect of
expert pruning guided by multimodal signals on language-only
capabilities.
\begin{itemize}
    \item \textbf{MMLU-Redux}~\cite{gema2025we}:
    A manually re-annotated subset of MMLU across 30 subjects with
    corrected or removed erroneous ground-truth labels, yielding a
    cleaner multi-domain knowledge benchmark. We use the corrected
    subset and follow the OpenCompass 0-shot no-CoT protocol.
    Metric: accuracy.

    \item \textbf{IFEval}~\cite{zhou2023ifeval}:
    Instruction-following benchmark composed of prompts with verifiable
    constraints.
    Metric: prompt-strict accuracy.

    \item \textbf{ARC Challenge}~\cite{clark2018arc}:
    Science exam questions requiring multi-step reasoning beyond simple
    retrieval.
    Metric: accuracy.
\end{itemize}

\textbf{Evaluation Pipeline.}
All benchmarks are evaluated using VLMEvalKit~\cite{duan2024vlmevalkit},
with the OpenCompass~\cite{2023opencompass} protocol for text-only
benchmarks.
For MCQ benchmarks (MMBench, MMStar, RealWorldQA, CV-Bench,
MMLU-Redux, ARC), we use \textsc{gpt-3.5-turbo-0125} as the
answer-extraction judge.
We evaluate the standard \textsc{Instruct} variants of
InternVL3.5-30B-A3B and Qwen3-VL-30B-A3B (the explicit
\textsc{Thinking} variants are \emph{not} used), so no model-level
chain-of-thought reasoning is engaged.
All pruned models---\method, HC-SMoE, and REAP---share the same
calibration set, and all models, including the unpruned baselines, are
evaluated under identical protocols.

\section{Additional Results}
\label{app:additional_results}

\subsection{Separate Multimodal and Text-Only Averages}
\label{app:separate_averages}

Table~\ref{tab:separate_averages} splits the ``Avg.''\ column of
Table~\ref{tab:main_results} into the mean retention over the eight
multimodal benchmarks and over the three text-only benchmarks.
At 50\%, \method attains the best multimodal average on all three
models; on text-only benchmarks it leads on InternVL3.5 and Gemma4 and
trails REAP on Qwen3-VL.
At 25\%, \method again has the best multimodal average on all three
models; where its text-only average trails, the gap is at most 1.7
points, and its 11-benchmark average remains best or tied on every
model.

\begin{table}[ht]
\centering
\small
\caption{Mean performance retention (\%) over the eight multimodal
benchmarks (MM), the three text-only benchmarks (Text), and all eleven
(All; the ``Avg.''\ column of Table~\ref{tab:main_results}).
Bold marks the best method per model, ratio, and column.}
\label{tab:separate_averages}
\begin{tabular}{@{}llcccc@{}}
\toprule
Model & Ratio & Method & MM & Text & All \\
\midrule
\multirow{6}{*}{InternVL3.5-30B-A3B}
 & \multirow{3}{*}{25\%} & HC-SMoE & 97.2 & 77.6 & 91.9 \\
 & & REAP & 97.9 & 90.5 & 95.9 \\
 & & \method & \textbf{98.6} & \textbf{95.3} & \textbf{97.7} \\
\cmidrule(lr){2-6}
 & \multirow{3}{*}{50\%} & HC-SMoE & 86.4 & 41.4 & 74.1 \\
 & & REAP & 91.6 & 80.3 & 88.5 \\
 & & \method & \textbf{94.0} & \textbf{83.8} & \textbf{91.2} \\
\midrule
\multirow{6}{*}{Qwen3-VL-30B-A3B}
 & \multirow{3}{*}{25\%} & HC-SMoE & 96.2 & \textbf{95.4} & 96.0 \\
 & & REAP & 94.7 & 95.1 & 94.8 \\
 & & \method & \textbf{97.5} & 94.0 & \textbf{96.6} \\
\cmidrule(lr){2-6}
 & \multirow{3}{*}{50\%} & HC-SMoE & 89.6 & 74.7 & 85.5 \\
 & & REAP & 88.4 & \textbf{81.8} & 86.6 \\
 & & \method & \textbf{92.2} & 77.9 & \textbf{88.3} \\
\midrule
\multirow{6}{*}{Gemma4-26B-A4B}
 & \multirow{3}{*}{25\%} & HC-SMoE & 74.7 & 83.8 & 77.2 \\
 & & REAP & 96.9 & \textbf{93.9} & \textbf{96.0} \\
 & & \method & \textbf{97.4} & 92.2 & \textbf{96.0} \\
\cmidrule(lr){2-6}
 & \multirow{3}{*}{50\%} & HC-SMoE & 38.9 & 40.6 & 39.3 \\
 & & REAP & 85.2 & 69.3 & 80.9 \\
 & & \method & \textbf{89.4} & \textbf{79.6} & \textbf{86.7} \\
\bottomrule
\end{tabular}
\end{table}

\subsection{Comparison with DiEP}
\label{app:diep}

DiEP~\cite{bai2025diep} learns a non-uniform, layer-wise expert budget
by differentiable search while scoring experts in a modality-agnostic
way; it therefore shares \method's allocation axis (global,
non-uniform) but not its scoring.
Table~\ref{tab:diep} compares the two on InternVL3.5-30B-A3B under the
identical calibration set, pipeline, and total budget (configuration
in Appendix~\ref{app:exp_settings}).
\method leads at both ratios and on nearly every benchmark, and the
gap widens with sparsity; DiEP's one win is MME at 50\% (the two are
essentially tied on MME at 25\%, and POPE separates no method).
DiEP's largest deficit is ChartQA, the benchmark most sensitive to
which experts survive (Section~\ref{sec:ablation_calibration}).
DiEP's learned allocation is far more depth-skewed than \method's
(averaged within depth quartiles it prunes 11/11/50/56 experts per
layer at 25\%, removing up to 69 of 128 experts from a single layer,
and leaves as few as 37 experts in a layer at 50\%), yet under the
same budget its retention falls short.
Combined with the decomposition in
Appendix~\ref{app:scoring_allocation}, this indicates that the
operative factor is which experts are selected rather than the
non-uniform budget itself.
DiEP additionally requires gradient-based search over its relaxation
parameters, whereas every \method score comes from a single forward
pass.

\begin{table}[ht]
\centering
\small
\caption{DiEP vs.\ \method on InternVL3.5-30B-A3B (each cell: 25\% /
50\% expert pruning; identical calibration set, pipeline, and judge).
MM-Ret.\ is the mean retention over these eight multimodal benchmarks.
HC-SMoE and REAP are reported in Table~\ref{tab:main_results}.}
\label{tab:diep}
\begin{tabular}{@{}lcc@{}}
\toprule
Benchmark & DiEP & \method (ours) \\
\midrule
MME          & 2364.3 / \textbf{2195.6} & \textbf{2364.9} / 2160.3 \\
MMBench      & 79.2 / 66.7 & \textbf{83.2} / \textbf{80.7} \\
MMStar       & 61.8 / 49.7 & \textbf{69.3} / \textbf{63.3} \\
ChartQA      & 54.0 / 40.4 & \textbf{87.0} / \textbf{76.7} \\
TextVQA      & 74.6 / 67.1 & \textbf{78.9} / \textbf{75.6} \\
RealWorldQA  & 70.6 / 64.7 & \textbf{71.6} / \textbf{71.6} \\
CV-Bench     & 81.3 / 69.4 & \textbf{84.3} / \textbf{83.3} \\
POPE         & 89.2 / 89.5 & \textbf{89.7} / \textbf{89.7} \\
\midrule
MM-Ret.\ (\%) & 90.6 / 79.9 & \textbf{98.6} / \textbf{94.0} \\
Parameter optimization & gradient search & none (single forward pass) \\
\bottomrule
\end{tabular}
\end{table}

\section{Modality Gap in Expert Importance}
\label{app:modality_gap}

\begin{figure}[!t]
\centering
\begin{minipage}[t]{0.49\textwidth}
\centering
\includegraphics[width=\linewidth]{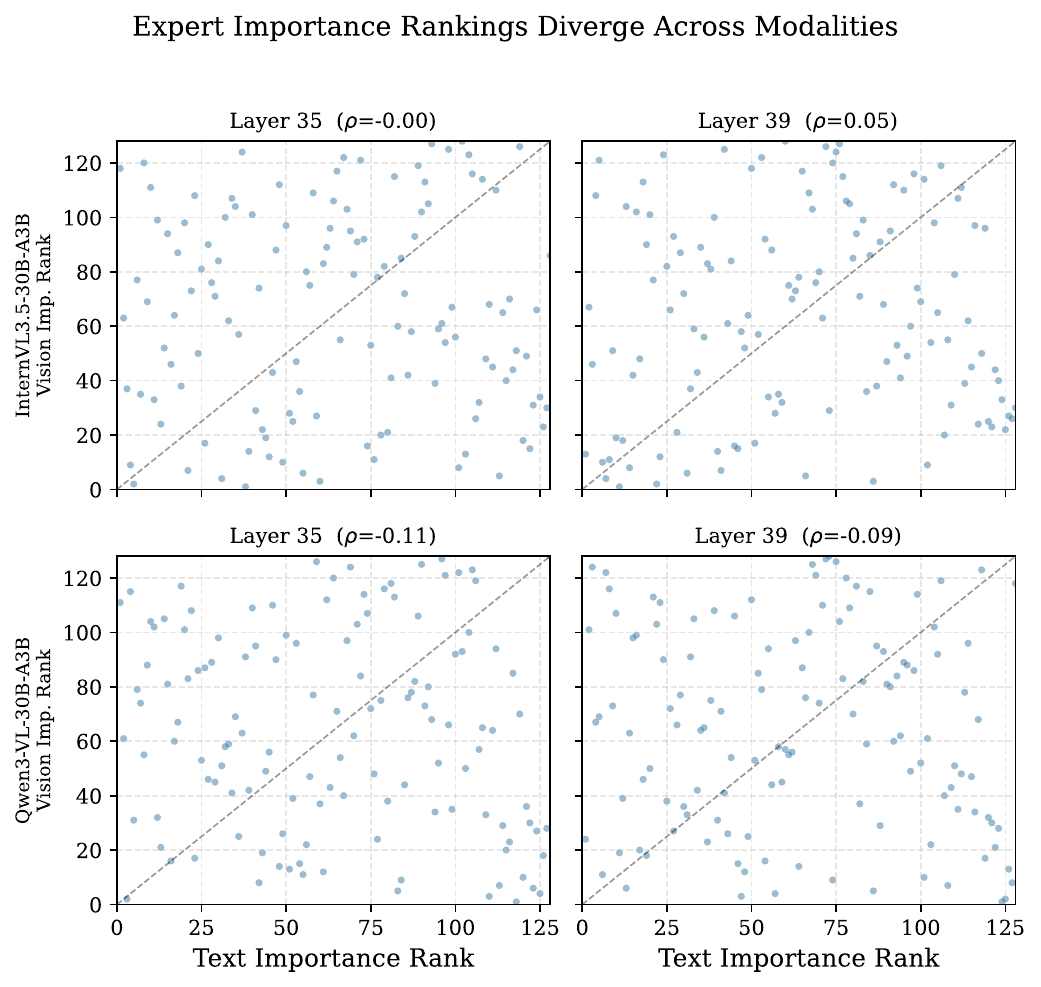}
\caption{Text vs.\ vision expert importance ranking ($\DeltaH$) for
InternVL3.5-30B-A3B and Qwen3-VL-30B-A3B at layers 35 and 39
(selected for their near-zero or negative Spearman $\rho$).
Each point is one expert; the dashed line indicates perfect
agreement.}
\label{fig:modality_gap}
\end{minipage}\hfill
\begin{minipage}[t]{0.49\textwidth}
\centering
\includegraphics[width=\linewidth]{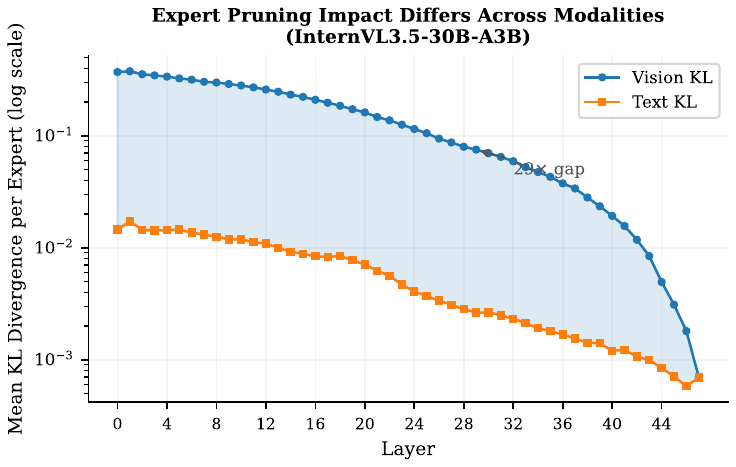}
\caption{Mean per-expert KL divergence at the LM-head output, on
vision-token (blue) vs.\ text-token (orange) positions, per layer for
InternVL3.5-30B-A3B. Each curve averages over experts removed one at
a time within the layer; the gap reaches up to $29\times$, indicating
that expert removal perturbs vision-position outputs substantially
more than text.}
\label{fig:vision_vs_text_kl}
\end{minipage}
\end{figure}

We provide empirical evidence that expert importance is
modality-dependent.
Figure~\ref{fig:modality_gap} plots text-side vs.\ vision-side
$\DeltaH$ for each expert at layers 35 and 39.
If experts served both modalities equally, all points would lie on the
diagonal; the observed scatter indicates that experts tend to
specialize by modality to varying degrees.

Figure~\ref{fig:vision_vs_text_kl} further quantifies this gap by
measuring the actual KL divergence at the LM head when each expert is
removed.
The vision-side KL (blue) consistently exceeds the text-side KL
(orange) across all layers, with the gap reaching $29\times$ in
shallow layers.

The per-model heterogeneity in
Table~\ref{tab:per_model_modality_gap} reflects differences in
modality-decomposed importance structure: Qwen3-VL exhibits
near-zero $\bar\rho_{\text{t},\text{v}}$ (rankings essentially
independent across modalities), while Gemma4 shows high
$\bar\rho_{\text{v},\text{la}}$ (vision and low-attention vision
rankings are strongly aligned), driving its small $\alpha$.

\begin{table}[h]
\centering
\small
\caption{Layer-averaged Spearman correlation between expert
importance rankings under three modality views (text, full vision,
low-attention vision) on the ShareGPT4V calibration set, computed
per layer and averaged across all MoE layers.
The closed-form expression $\alpha = a/(a+b)$ with
$a := \bar\rho_{\text{t},\text{v}} - \bar\rho_{\text{t},\text{la}}$
and $b := \bar\rho_{\text{v},\text{la}}$ recovers the
per-model values used in the main results.}
\label{tab:per_model_modality_gap}
\begin{tabular}{lcccc}
\toprule
Model & $\bar\rho_{\text{t},\text{v}}$
& $\bar\rho_{\text{t},\text{la}}$
& $\bar\rho_{\text{v},\text{la}}$
& closed-form $\alpha$ \\
\midrule
InternVL3.5-30B-A3B          & $+0.38$ & $-0.27$ & $+0.22$ & $0.75$ \\
Qwen3-VL-30B-A3B-Instruct    & $+0.01$ & $-0.22$ & $+0.46$ & $0.33$ \\
Gemma4-26B-A4B-Instruct      & $+0.63$ & $+0.55$ & $+0.91$ & $0.08$ \\
\bottomrule
\end{tabular}
\end{table}
This modality gap motivates decomposing importance by modality in
our combined scoring (Eq.~\ref{eq:combined_score}), where text-side
$\DeltaH$ acts as a preservation signal and low-attention vision-side
$\DeltaH$ acts as a pruning signal (Section~\ref{sec:low_attn}; the
sign convention is verified in Appendix~\ref{app:kl_sign}).

\section{Cross-Dataset Stability of Expert Rankings}
\label{app:low_attn_analysis}

\begin{figure}[!ht]
\centering
\includegraphics[width=0.9\textwidth]{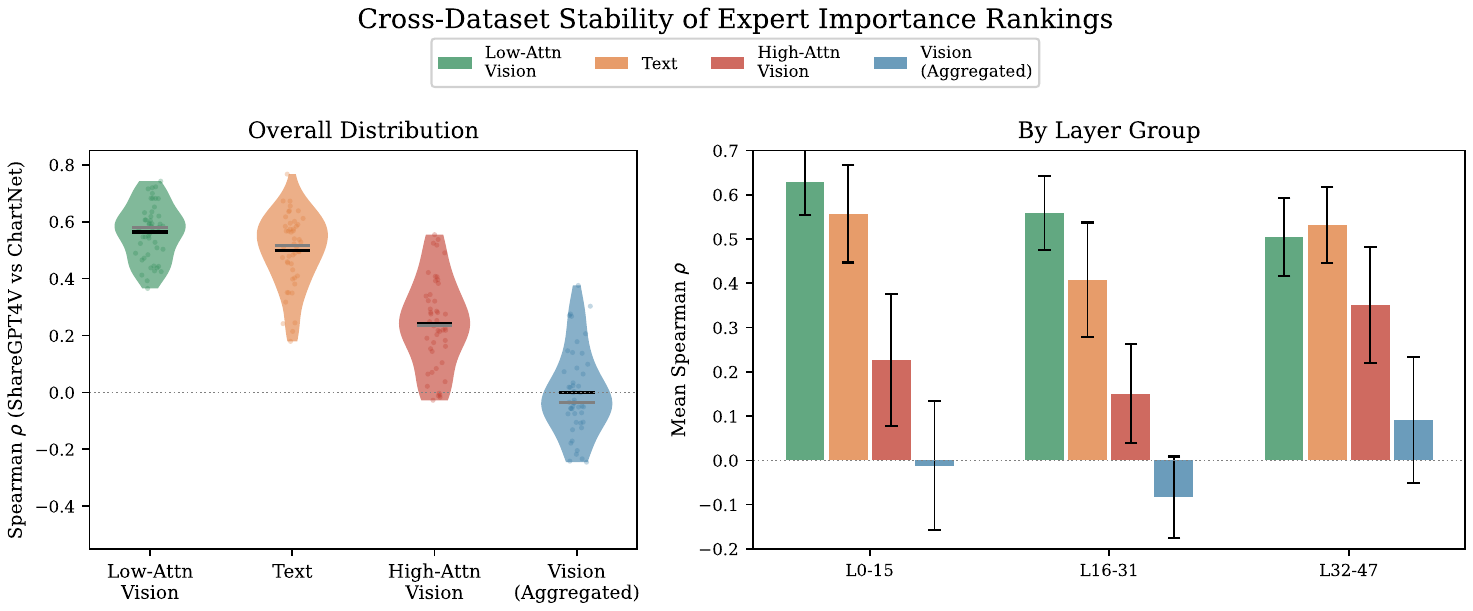}
\caption{Cross-dataset stability of expert importance rankings
(Spearman $\rho$ between ShareGPT4V and ChartNet calibration sets)
for InternVL3.5-30B-A3B.
Left: overall distribution of per-layer $\rho$ by token type (violins
with mean and median markers).
Right: mean $\rho$ by layer group (L0--15, L16--31, L32--47).
Low-attention vision tokens yield the most stable rankings across
calibration sources ($\bar{\rho}{\approx}0.56$), substantially
exceeding text ($\approx 0.50$), high-attention vision
($\approx 0.24$ at $q{=}20\%$; the matched-size $q{=}2\%$ value
reported in \S\ref{sec:ablation_components} is higher),
and full vision ($\approx 0$); this supports their
use as the vision-side pruning signal.}
\label{fig:calibration_stability}
\end{figure}

A key design choice in \method is to use low-attention vision tokens
rather than all vision tokens or high-attention ones.
Figure~\ref{fig:calibration_stability} validates this choice by
measuring the cross-dataset stability of expert importance rankings:
we compute $\DeltaH$-based rankings using ShareGPT4V and ChartNet as
separate calibration sets and report the Spearman $\rho$ between them.
The chart-domain source (ChartNet) is disjoint from the evaluation
suite and matches the setup used in
Section~\ref{sec:ablation_calibration} and
Figure~\ref{fig:quantile_ablation}.

\subsection{Alternative Vision-Token Selectors}
\label{app:vision_selectors}

Since expert pruning operates in the MoE blocks, attention is not the
only conceivable signal for selecting the visual calibration tokens.
Table~\ref{tab:vision_selectors} screens seven selectors on the two
properties a calibration signal needs: stability of the induced expert
ranking across calibration datasets ($\rho_{\text{stab}}$; ShareGPT4V
vs.\ ChartQA images, the latter used only as a held-out image set and
not for calibrating any pruned model in this paper) and distinctness
from the text signal ($\rho_{\text{dist}}$; lower is more distinct).
Besides the three attention-based signals, we include representation
drift (one minus the cosine similarity between a token's input and
output representations across the MoE layers; keeping the
highest-drift tokens is also the selection rule of Li et
al.~\cite{li2026atvpruning}), diversity (farthest-point sampling), and
random selection.
For tractability the screening uses routing-mass importance rather
than the exact $\DeltaH$; it reproduces the $\DeltaH$-based ordering of
Figure~\ref{fig:calibration_stability} for the attention-based
signals, so only the relative comparison is used.

Low-attention is the only selector that is both the most stable and
the most distinct from the text signal; full-vision and random are
unstable, drift-low is only weakly stable, and high-attention,
drift-high, and diversity produce text-like importance.
Run end-to-end through the full $\DeltaH$ calibration (InternVL3.5,
50\%; only the vision-token subset changed), the drift-high rule
selects a very different expert set (Jaccard $0.33$ against
\method's) and loses on all four ablation benchmarks
(\method $\rightarrow$ drift-high: MME $2160\rightarrow1490$, MMStar
$63.3\rightarrow49.9$, TextVQA $75.6\rightarrow41.9$, ChartQA
$76.7\rightarrow71.5$).
Attention received from the answer-generating text stream measures
whether a token's content reaches the output, which is what pruning
must decide; drift and diversity instead measure how strongly a token
is transformed or how it spreads in representation space.

\begin{table}[ht]
\centering
\small
\caption{Vision-token selector screening on InternVL3.5-30B-A3B
(routing-mass importance; ShareGPT4V vs.\ ChartQA images as a held-out set, 60 samples each;
every subset selector keeps 2\% of the vision tokens).
Higher $\rho_{\text{stab}}$ is better; lower $\rho_{\text{dist}}$ means
more distinct from the text signal.}
\label{tab:vision_selectors}
\begin{tabular}{@{}lcc@{}}
\toprule
Selector & $\rho_{\text{stab}}$ (stability) & $\rho_{\text{dist}}$ (vs.\ text; low = distinct) \\
\midrule
Low-attention (ours)       & \textbf{0.51} & $\mathbf{-0.12}$ \\
High-attention             & 0.30 & 0.47 \\
Full vision (no subset)    & $-0.23$ & 0.30 \\
Drift-low                  & 0.19 & 0.01 \\
Drift-high                 & 0.29 & 0.34 \\
Diversity (farthest-point) & 0.36 & 0.36 \\
Random (3 seeds)           & $-0.23$ & 0.30 \\
\midrule
Text signal (reference)    & 0.30 & --- \\
\bottomrule
\end{tabular}
\end{table}

\subsection{Spatial Location of Low-Attention Tokens}
\label{app:spatial_stats}

To verify that bottom-2\% low-attention tokens correspond to
non-salient content, we mapped them onto the whole-image thumbnail
grid of InternVL3.5-30B-A3B (ShareGPT4V calibration images; 60 images
$\times$ 48 layers; per-layer bottom 2\%) and compared them with a
spectral-residual saliency map of each image
(Table~\ref{tab:spatial_stats}).
Low-attention tokens fall overwhelmingly outside salient regions: a
bottom-2\% token is about $2.9\times$ less likely than an average
vision token to lie in the top-50\% saliency-mass region, and about
$2.8\times$ less likely to lie in the Otsu-thresholded foreground.
A purely geometric border measure is uninformative here, because the
image border attracts the \emph{highest}-attention tokens (top-2\%
tokens are about $1.8\times$ over-represented on the border ring, a
fixed-position pattern consistent with attention-sink behavior),
whereas bottom-2\% tokens are under-represented there ($0.69\times$ the
area expectation).

\begin{table}[ht]
\centering
\small
\caption{Saliency statistics of bottom-2\% low-attention vision tokens
(InternVL3.5-30B-A3B, ShareGPT4V calibration; 60 images $\times$ 48
layers).
Base rate: the same statistic over all vision tokens.}
\label{tab:spatial_stats}
\begin{tabular}{@{}lcc@{}}
\toprule
Statistic & Low-attention tokens & Base rate \\
\midrule
Fraction outside the top-50\% saliency mass & 92.0\% & 76.5\% \\
Fraction in the saliency-map background (Otsu threshold) & 89.5\% & 70.1\% \\
\bottomrule
\end{tabular}
\end{table}

\subsection{Which Signal Carries Calibration-Domain Sensitivity}
\label{app:calib_swap}

Table~\ref{tab:ablation}(b) shows that changing the calibration domain
changes the prune decisions.
To locate where this sensitivity enters, we swapped the calibration
data of each signal separately on Qwen3-VL at 50\% pruning and compared
the resulting prune map with the original ShareGPT4V map
(Table~\ref{tab:calib_swap}).
Swapping only the text-side calibration to the ChartNet mixture moves
the prune map about $3\times$ more than swapping only the vision-side
signal (as the ratio of decision-flip rates) and reproduces most of the
mask movement of the full mixture: of the experts whose prune/keep
decision flips under the full mixture, 82\% also flip under the
text-only swap, vs.\ 13\% under the vision-only swap.
Domain composition is thus carried by the text-side preservation
signal, while the low-attention vision signal stays stable, consistent
with Figure~\ref{fig:calibration_stability}; the practical guidance
that follows is given in Appendix~\ref{app:limitations}.

\begin{table}[ht]
\centering
\small
\caption{$2{\times}2$ calibration swap on Qwen3-VL-30B-A3B (50\%
pruning): Jaccard similarity of the resulting prune map to the original
ShareGPT4V map, and the number of prune/keep decisions that flip (of
6{,}144 experts; each rescued expert pairs with a newly pruned one).}
\label{tab:calib_swap}
\begin{tabular}{@{}lcc@{}}
\toprule
Calibration swapped to the ChartNet mixture & Jaccard vs.\ original & Decisions flipped \\
\midrule
Vision (low-attention) side only & 0.922 & 248 \\
Text side only                   & 0.776 & 776 \\
Both (full mixture; Table~\ref{tab:ablation}(b)) & 0.780 & 760 \\
\bottomrule
\end{tabular}
\end{table}

\section{Validation of Fixed Top-\texorpdfstring{$k$}{k} Counterfactual}
\label{app:topk_counterfactual}

Our importance metric (Eq.~\ref{eq:deltah}) uses a fixed top-$k$
counterfactual that does not evaluate the promoted $(k{+}1)$-th expert,
whose output is set to zero.
An alternative is to evaluate the promoted $(k{+}1)$-th expert so that
exactly $k$ experts remain active, matching the router's actual behavior
in the pruned model.
We compare the two counterfactuals on InternVL3.5-30B-A3B by computing
$\DeltaH$ under each variant for every expert in every layer.

Table~\ref{tab:topk_counterfactual} shows that the two counterfactuals
produce nearly identical rankings and pruning decisions across all
modalities. We therefore adopt the simpler fixed top-$k$ variant,
which avoids evaluating the promoted $(k{+}1)$-th expert at negligible
cost in ranking fidelity or prune-mask selection.

\begin{table}[h]
\centering
\scriptsize
\setlength{\tabcolsep}{6pt}
\caption{Agreement between the fixed top-$k$ and promote-$(k{+}1)$
counterfactuals on InternVL3.5-30B-A3B. Spearman $\rho$ is the mean
across 48 layers; Jaccard is computed at $50\%$ per-layer pruning of
the lowest-score experts.}
\label{tab:topk_counterfactual}
\begin{tabular}{l c c}
\toprule
Modality & Spearman $\rho$ & Jaccard @ $50\%$ \\
\midrule
Text             & 0.999 & 0.97 \\
Low-attn vision  & 0.987 & 0.92 \\
High-attn vision & 0.994 & 0.93 \\
Full vision      & 0.981 & 0.89 \\
\bottomrule
\end{tabular}
\end{table}

\section{Validating \texorpdfstring{$\DeltaH$}{Delta-H} as an Importance Proxy}
\label{app:spearman_depth}

\begin{figure}[t]
\centering
\includegraphics[width=0.85\textwidth]{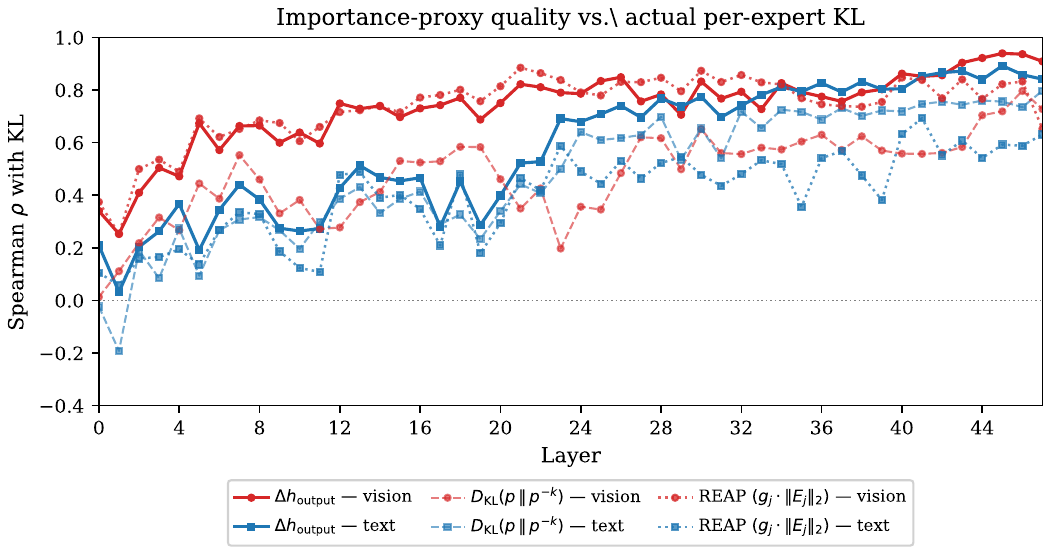}
\caption{Per-layer Spearman $\rho$ between importance proxies and
actual KL divergence for InternVL3.5-30B-A3B.
$\DeltaH$ (solid) is the strongest proxy on text tokens, substantially
exceeding both routing-KL (dashed) and REAP's activation-norm
product (dotted); on vision tokens, $\DeltaH$ and REAP are comparable,
while routing-KL trails both.
For vision tokens, $\DeltaH$ correlation exceeds $0.8$ in deep layers
(L36--L47); for text tokens, $\DeltaH$ rises from ${\sim}0.05$ in
shallow layers to ${\sim}0.89$ in deep layers, suggesting that
$\DeltaH$ is a tighter proxy where it matters most.}
\label{fig:spearman_depth}
\end{figure}

To validate $\DeltaH$ as an importance proxy, we compare it against
two baselines. The first is the \textbf{routing-KL} when expert $j$
is masked---a principled distributional distance that quantifies how
much the router distribution shifts when expert $j$ is removed:
\begin{equation}
\Delta\mathrm{KL}_{\text{routing}}^{(l,j)} = \mathbb{E}_{x}\Big[
  D_{\mathrm{KL}}\!\big(p^{(l)}(x)\,\|\,p^{(l,-j)}(x)\big)
\Big],
\end{equation}
where $p^{(l,-j)}$ is obtained by zeroing $p_j$ and renormalizing
the remaining entries, matching the redistribution used in
$\DeltaH$ (we additionally apply Laplace smoothing with
$\epsilon_{\text{KL}}{=}10^{-12}$ to prevent numerical infinities without
affecting rank ordering).
A larger $\Delta\mathrm{KL}_{\text{routing}}$ indicates that
expert $j$ carries a more distinctive routing pattern.
The second is \textbf{REAP}'s~\cite{lasby2025reap} router-weighted
activation norm, which captures each expert's magnitude contribution:
\begin{equation}
s_{\text{REAP}}^{(l,j)} = \mathbb{E}_{x}\big[\,
  p_j^{(l)}(x) \cdot \|E_j^{(l)}(x)\|_2
\,\big],
\end{equation}
where $p_j^{(l)}(x)$ is the post-softmax router gate value (defined
in Section~\ref{sec:prelim_moe}) and $E_j^{(l)}(x)$ is expert $j$'s
output.
Figure~\ref{fig:spearman_depth} reports the per-layer Spearman rank
correlation between each of these proxies and the actual KL
divergence at the LM head caused by expert removal.
For computational tractability, the ground-truth KL is averaged over
$n{=}10$ calibration samples per (layer, expert) pair, since exact
computation requires re-running the post-MoE network once per ablated
expert.

\subsection{Causal Check of the Vision-Side Sign Convention}
\label{app:kl_sign}

Section~\ref{sec:low_attn} treats a high $\Delta H_{\text{la}}$ as
evidence for removal.
To verify this sign direction at the model output, we isolated the
low-attention signal ($\alpha{=}1$, 25\% pruning,
InternVL3.5-30B-A3B), ranked experts by $\Delta H_{\text{la}}$, and
compared three removal orders, measuring the per-token LM-head KL
divergence each induces against the unpruned model (same protocol as
above; $n{=}10$ calibration samples, 3{,}948 text and 17{,}152 vision
positions).
Table~\ref{tab:kl_sign} shows that removing the
highest-$\Delta H_{\text{la}}$ experts first perturbs the
output-generating text positions least, random removal is
intermediate, and the inverted order perturbs them most, which
confirms that high-$\Delta H_{\text{la}}$ experts contribute least to
the output distribution.

\begin{table}[ht]
\centering
\small
\caption{Mean per-token LM-head KL divergence against the unpruned
model at the output-generating (text) positions, by expert-removal
order on InternVL3.5-30B-A3B ($\alpha{=}1$; 25\% of experts removed
from the indicated end of the $\Delta H_{\text{la}}$ ranking;
$n{=}10$).
At vision positions, which generate no tokens and are perturbed
strongly by any removal (Figure~\ref{fig:vision_vs_text_kl}), the
three orders give 1.24, 1.01, and 1.30; that our order exceeds random
removal there shows that the removed experts are not inert.}
\label{tab:kl_sign}
\begin{tabular}{@{}lc@{}}
\toprule
Removal order & KL at text positions \\
\midrule
Highest-$\Delta H_{\text{la}}$ experts first (our sign) & \textbf{0.30} \\
Randomly selected experts & 0.48 \\
Lowest-$\Delta H_{\text{la}}$ experts first (inverted) & 1.13 \\
\bottomrule
\end{tabular}
\end{table}

\section{Cross-Layer Magnitude Variation of \texorpdfstring{$\DeltaH$}{Delta-H}}
\label{app:deltah_dist}

Raw $\DeltaH$ values exhibit two properties that necessitate the
log-transform and layer-wise z-score normalization described in
Section~\ref{sec:global_pruning}.

\begin{figure}[t]
\centering
\includegraphics[width=\textwidth]{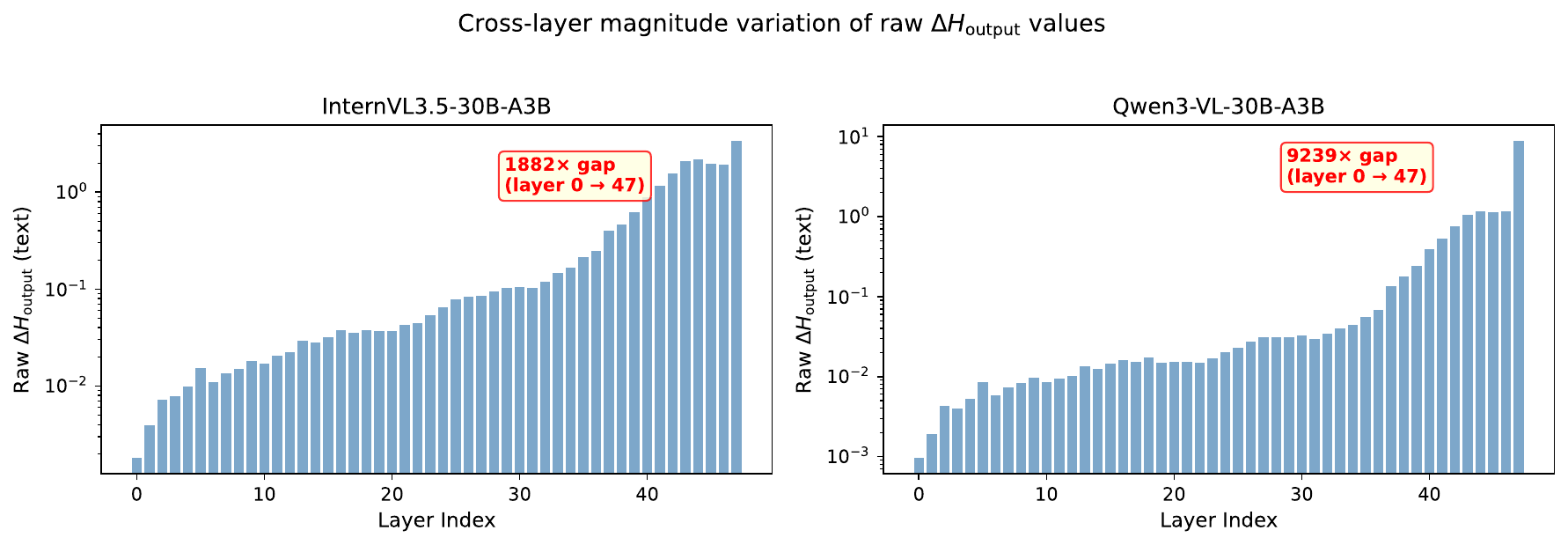}
\caption{Mean $\DeltaH$ (text tokens) per layer for InternVL3.5-30B-A3B
and Qwen3-VL-30B-A3B. Y-axis is log-scaled. Deep layers produce
values $10^3$--$10^4\times$ larger than shallow layers, necessitating
layer-wise normalization before cross-layer comparison.}
\label{fig:deltah_dist}
\end{figure}

\textbf{Cross-layer magnitude variation.}
Figure~\ref{fig:deltah_dist} shows the mean $\DeltaH$ per layer for
InternVL3.5-30B-A3B and Qwen3-VL-30B-A3B on the ShareGPT4V calibration set.
The deepest layer exceeds the shallowest by
${\sim}1{,}900\times$ (InternVL) and ${\sim}9{,}200\times$ (Qwen),
spanning nearly four orders of magnitude.
Without layer-wise normalization, global ranking would be dominated by
deep-layer experts regardless of their relative importance within
each layer.

\begin{figure}[t]
\centering
\includegraphics[width=\textwidth]{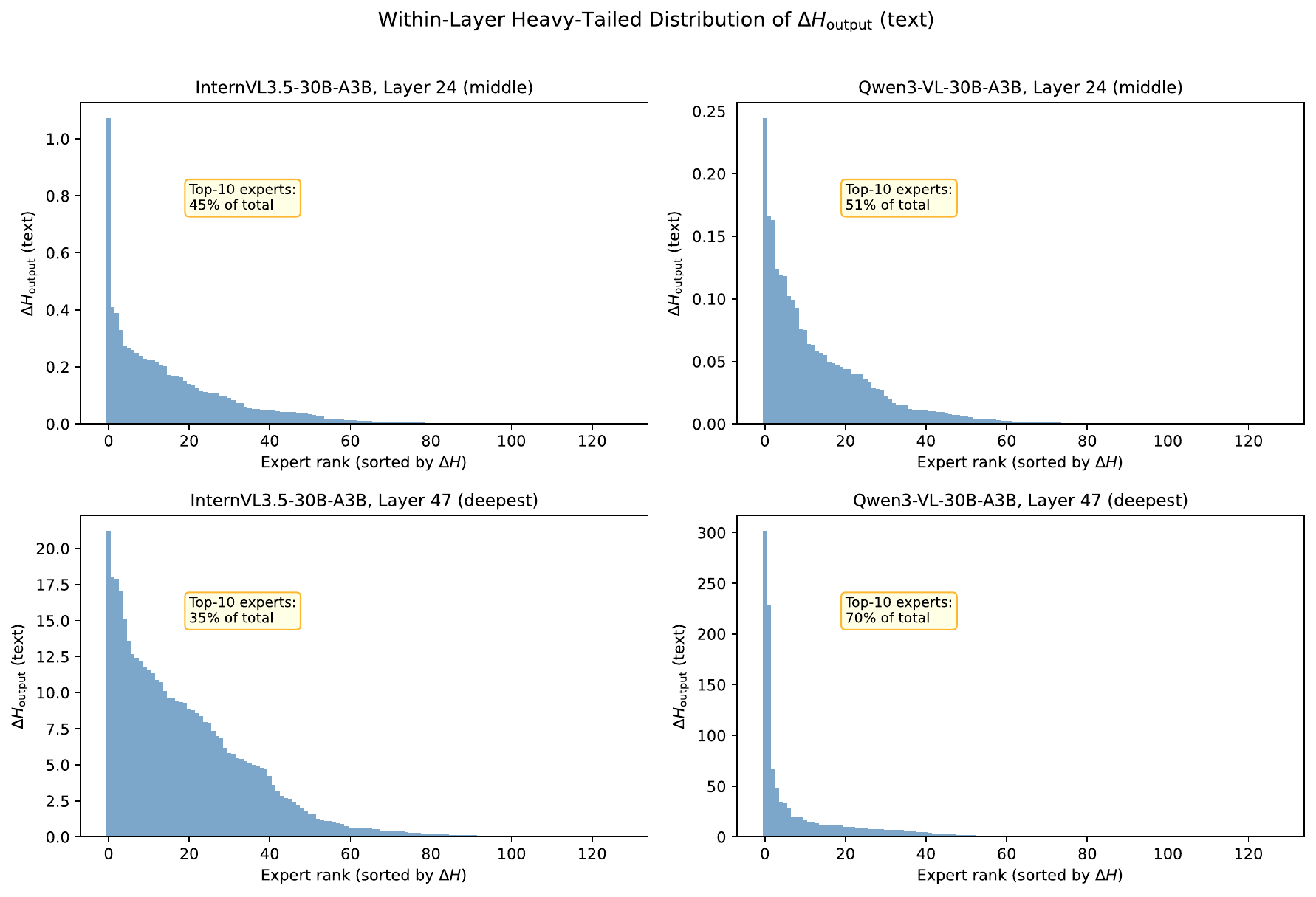}
\caption{Within-layer distribution of $\DeltaH$ (text tokens) for
selected layers of InternVL3.5-30B-A3B and Qwen3-VL-30B-A3B.
Expert importance follows a heavy-tailed distribution, with the
top-10 experts contributing 35--70\% of total $\DeltaH$ per layer.}
\label{fig:deltah_longtail}
\end{figure}

\textbf{Within-layer heavy-tailed distribution.}
Figure~\ref{fig:deltah_longtail} shows the per-expert $\DeltaH$
distribution within representative layers, sorted in descending order.
In both models, a small number of experts account for a
disproportionate share of the total importance
(e.g., the top-10 experts contribute 35--70\% of the layer's total
$\DeltaH$).
The log-transform compresses this heavy tail, ensuring that the
z-score normalization produces well-separated scores across all experts
rather than being dominated by a few outliers.

\section{Per-Layer Pruning Distribution under Global Pruning}
\label{app:pruning_dist}

\begin{figure}[t]
\centering
\includegraphics[width=\textwidth]{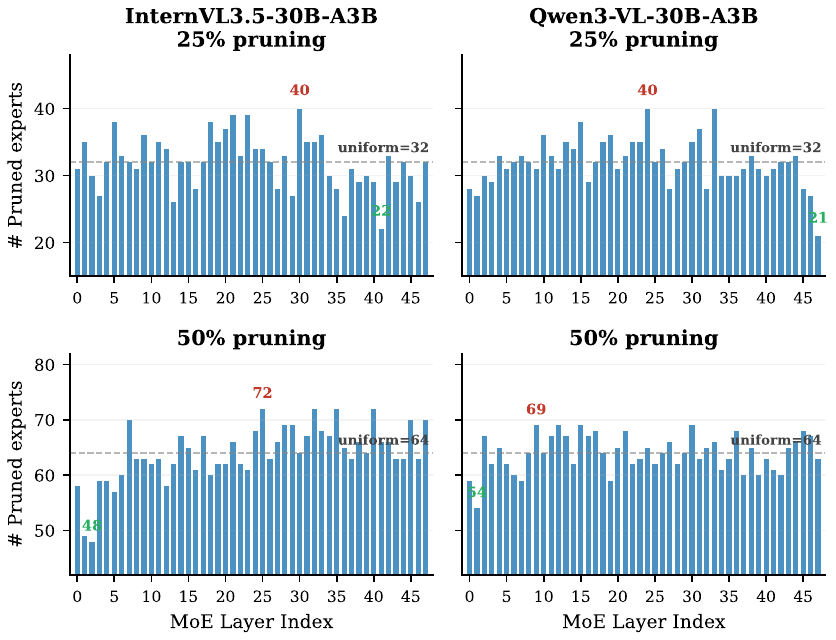}
\caption{Per-layer pruned expert count under global pruning for
InternVL3.5-30B-A3B and Qwen3-VL-30B-A3B at $25\%$ and $50\%$ pruning
ratios. Gray dashed lines indicate the uniform per-layer baseline
(32 and 64 experts, respectively); red and green annotations mark
the layers with the maximum and minimum pruned-expert counts.}
\label{fig:pruning_dist}
\end{figure}

We highlight three observations from
Figure~\ref{fig:pruning_dist}.
(1)~Global pruning distributes the budget adaptively---some layers
contribute more pruning candidates than others, rather than enforcing
a fixed quota.
(2)~Although our algorithm does not enforce a per-layer minimum, no
layer is stripped below the top-$k$ threshold ($k{=}8$) in practice:
the minimum number of remaining experts across all layers is 56 for InternVL
and 59 for Qwen at 50\% pruning, ensuring that sparse routing remains
meaningful. At higher sparsities where a layer might lose all
experts, this would correspond to our score identifying that
layer's MoE as redundant; the layer naturally falls back to its
residual identity path (still a valid transformer block), rather
than constituting a routing failure.
(3)~In Qwen3-VL, the final layers (L44--L47) are pruned less
aggressively under 25\% pruning, reflecting a more concentrated
$\tilde z$ distribution in deep layers (\S\ref{sec:ablation_components}).

\section{Scoring vs.\ Allocation and \texorpdfstring{$\alpha$}{alpha} under a Uniform Budget}
\label{app:scoring_allocation}

\method differs from the uniform baselines in
Table~\ref{tab:main_results} along two axes: the modality-aware score
and the global, non-uniform allocation.
Table~\ref{tab:scoring_allocation} crosses the two on
InternVL3.5-30B-A3B at 50\% pruning.
Applying REAP's scores under a global budget changes essentially
nothing: after the same per-layer log$+$z-score normalization, its
scores induce a near-uniform allocation (the global and uniform REAP
masks agree at Jaccard $0.955$; 140 of the 6{,}144 prune/keep decisions
flip), and performance is unchanged within noise.
Within the global column, the modality-aware score is ahead on all
four benchmarks and decisively so on ChartQA.
The DiEP comparison (Appendix~\ref{app:diep}) holds the allocation
axis fixed and varies only the scoring, so the two analyses isolate
the scoring contribution from both directions.

\textbf{Why the uniform variant needs a different $\alpha$.}
A uniform per-layer quota uses the composite score of
Eq.~\ref{eq:combined_score} as a within-layer ordering.
That score is a removability score whose terms are weighted for the
global operating point: on InternVL3.5, text-side and low-attention
importance are anti-correlated (Table~\ref{tab:per_model_modality_gap}), so removing
the highest-$\Delta H_{\text{la}}$ experts already steers clear of
text-critical ones, and the global cut stops early enough for this to
hold.
A forced per-layer 50\% cut instead runs past the experts the
low-attention signal clearly flags; from there, protection falls to
the text term, whose weight $1{-}\alpha$ is only $0.2$ at
$\alpha{=}0.8$, too little to keep text-critical experts out of the
cut.
The mechanism predicts that a uniform budget requires a smaller
$\alpha$, which Table~\ref{tab:alpha_uniform} confirms: at
$\alpha{=}0.5$, uniform \method is at least competitive with uniform
REAP on every benchmark (within about one point on MME and MMStar,
ahead on ChartQA and TextVQA); TextVQA stays flat and high across the
whole band $\alpha\le0.5$, ChartQA improves with more vision-side
evidence up to $\alpha{=}0.5$, and $\alpha\ge0.7$ is unstable under a
uniform quota.
The recovery is not a convergence to REAP's selection: the re-tuned
masks remain far from REAP's (Jaccard $0.47$--$0.49$ across the band).
We therefore report the re-tuned $\alpha{=}0.5$ variant as the uniform
baseline in Table~\ref{tab:ablation}(a).
Global allocation still leads the re-tuned uniform variant clearly on
MME and ChartQA (MMStar and TextVQA are within about one point either
way), so we retain the global operating point.

\begin{table}[ht]
\centering
\small
\caption{Scoring $\times$ allocation on InternVL3.5-30B-A3B at 50\%
pruning (MME / MMStar / ChartQA / TextVQA).}
\label{tab:scoring_allocation}
\resizebox{\textwidth}{!}{%
\begin{tabular}{@{}lcc@{}}
\toprule
 & Uniform per-layer budget & Global non-uniform budget \\
\midrule
REAP scoring    & 2109.6 / 63.3 / 65.3 / 74.4 (Table~\ref{tab:main_results}) & 2085.7 / 63.0 / 65.1 / 74.6 \\
\method scoring & 2103.9 / 62.2 / 72.1 / 76.5 ($\alpha{=}0.5$; Table~\ref{tab:ablation}(a)) & \textbf{2160.3} / \textbf{63.3} / \textbf{76.7} / 75.6 (Table~\ref{tab:main_results}) \\
\bottomrule
\end{tabular}}
\end{table}

\begin{table}[ht]
\centering
\small
\caption{Uniform per-layer 50\% pruning on InternVL3.5-30B-A3B with
$\alpha$ re-tuned for the uniform budget.
All rows were evaluated with the same, final version of our
evaluation pipeline.}
\label{tab:alpha_uniform}
\begin{tabular}{@{}lcccc@{}}
\toprule
Score (uniform per-layer budget) & MME & MMStar & ChartQA & TextVQA \\
\midrule
\method, $\alpha{=}0.0$ & 2144.1 & 60.1 & 56.2 & 76.5 \\
\method, $\alpha{=}0.1$ & 2155.9 & 59.8 & 56.8 & 76.8 \\
\method, $\alpha{=}0.2$ & 2138.0 & 61.4 & 60.9 & 76.9 \\
\method, $\alpha{=}0.3$ & 2124.6 & 60.9 & 61.2 & 76.6 \\
\method, $\alpha{=}0.5$ (Table~\ref{tab:ablation}(a)) & 2103.9 & 62.2 & 72.1 & 76.5 \\
\method, $\alpha{=}0.7$ & 1675.4 & 49.3 & 44.8 & 58.2 \\
\method, $\alpha{=}0.8$ (global operating point) & 1918.4 & 52.8 & 61.6 & 61.2 \\
\midrule
REAP (uniform; Table~\ref{tab:main_results}) & 2109.6 & 63.3 & 65.3 & 74.4 \\
\bottomrule
\end{tabular}
\end{table}

\section{Qualitative Examples}
\label{app:qualitative}

\begin{figure}[t]
\centering
\begin{minipage}[t]{0.22\textwidth}
\vspace{0pt}
\includegraphics[width=\linewidth]{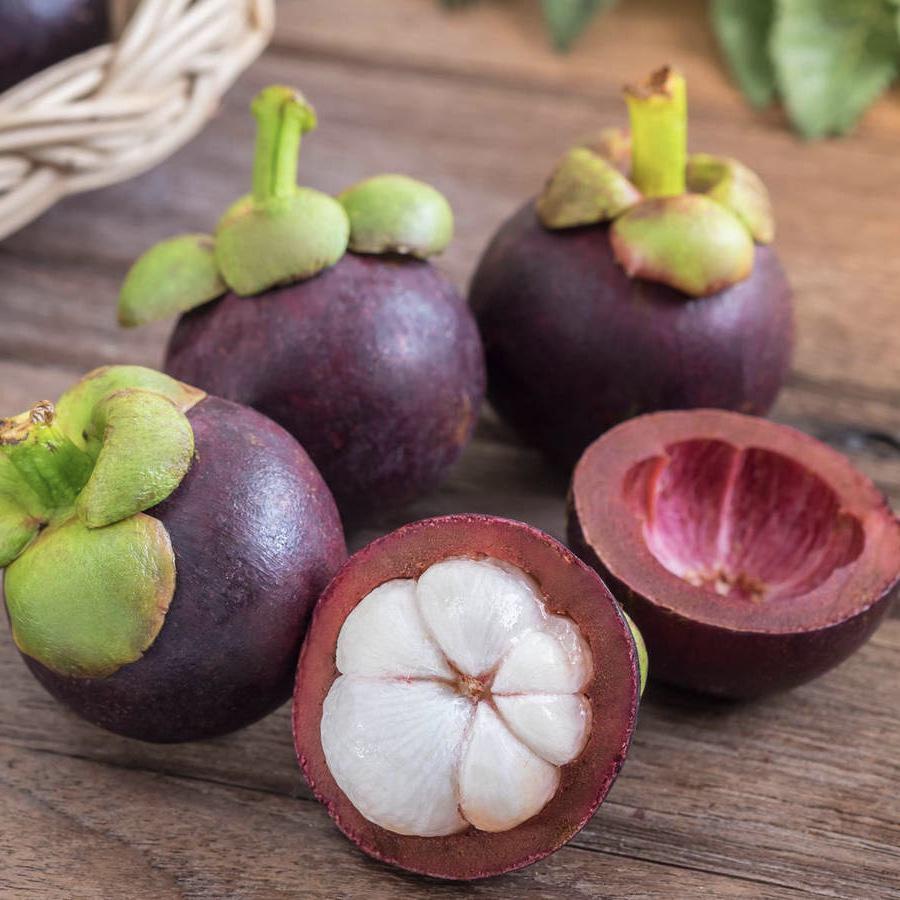}
\end{minipage}\hfill%
\begin{minipage}[t]{0.75\textwidth}\scriptsize\vspace{0pt}
\textbf{Q:} How many uncut fruits are in the image?\\[0.3em]
\textbf{Unpruned:} \textcolor{red}{\textbf{Four uncut fruits.}} A
detailed breakdown: \textcolor{red}{top-left, top-center, top-right,
and bottom-left}---each whole, dark-purple with green leaves;
\textcolor{red}{the remaining two are cut open and excluded.}\\[0.3em]
\textbf{HC-SMoE:} \textcolor{red}{In the image, there are five uncut
fruits.}\\[0.3em]
\textbf{\method (50\%):} \textbf{Three whole mangosteens with intact
skins, plus one cut in half} revealing the white flesh inside; brief
reasoning over each fruit's state.
\end{minipage}
\par\vspace{0.5em}\hrule\vspace{0.5em}
\begin{minipage}[t]{0.22\textwidth}\vspace{0pt}
\includegraphics[width=\linewidth]{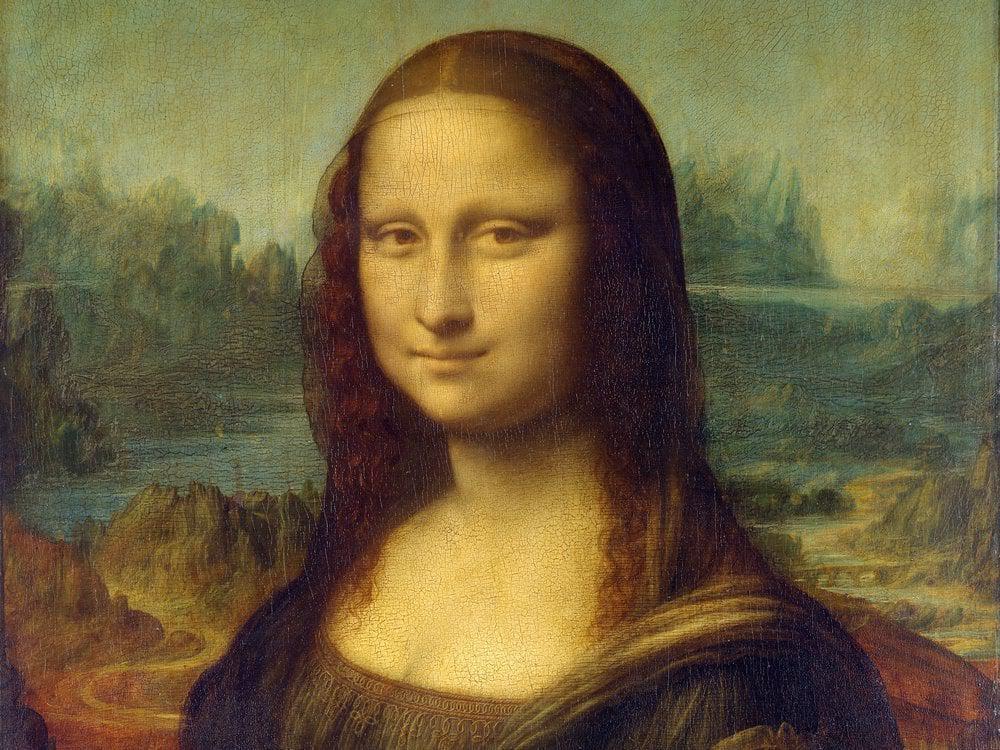}
\end{minipage}\hfill%
\begin{minipage}[t]{0.75\textwidth}\scriptsize\vspace{0pt}
\textbf{Q:} Do you know who paint this?\\[0.3em]
\textbf{Unpruned:} The painting is the \textbf{Mona Lisa}, by
\textbf{Leonardo da Vinci} (1452--1519). Discusses sfumato technique,
the High-Renaissance period (1503--1517), and the work's location at
the Louvre.\\[0.3em]
\textbf{HC-SMoE:} \textcolor{red}{The painting is created by the
artist known as ``InternVL''. This painting is a part of a multi-modal
language model.}\\[0.3em]
\textbf{\method (50\%):} The painting is the \textbf{Mona Lisa}, by
\textbf{Leonardo da Vinci}. Mentions sfumato (mis-spelled
``sfonto''), the High Renaissance, and the Louvre Museum.
\end{minipage}
\caption{Qualitative comparisons on LLaVA-Bench (in-the-wild) for
InternVL3.5-30B-A3B at $50\%$ pruning. Selected as the two samples
with the most pronounced output divergence across configurations.
\textcolor{red}{Red} marks factually incorrect or hallucinated
content; \textbf{bold} marks key factual content.}
\label{fig:qualitative}
\end{figure}

Figure~\ref{fig:qualitative} contrasts model outputs across three
conditions on LLaVA-Bench (in-the-wild)~\cite{liu2023llava}: the
unpruned baseline, HC-SMoE at 50\% pruning, and \method at 50\%
pruning.
In the counting case (top), \method correctly identifies the fruit
composition (three whole mangosteens and one halved fruit), whereas
HC-SMoE substantially overcounts and even the unpruned baseline
miscounts.
In the Mona Lisa case (bottom), \method matches the unpruned baseline
in correctly attributing the painting and its art-historical context,
while HC-SMoE produces a self-referential hallucination of model
identity.

\end{document}